\documentclass[lettersize,journal]{IEEEtran}
\usepackage{amsmath,amsfonts}
\usepackage{bm}
\usepackage{algorithmic}
\usepackage{algorithm}
\usepackage{array}
\usepackage{subcaption}
\usepackage{textcomp}
\usepackage{stfloats}
\usepackage{url}
\usepackage{verbatim}
\usepackage{graphicx}
\usepackage{cite}
\usepackage{booktabs}
\usepackage{multirow}
\usepackage{todonotes}
\usepackage[hidelinks]{hyperref}
\newcommand{\etal}{~\textit{et~al}.}
\DeclareMathOperator{\ntfF}{\mathcal{F}}
\DeclareMathOperator{\ntfV}{\mathcal{V}}
\DeclareMathOperator{\ntfS}{\mathcal{S}}
\DeclareMathOperator{\ntfSL}{\mathcal{S}_{\text{L}}}
\DeclareMathOperator{\ntfSH}{\mathcal{S}_{\text{H}}}

 \DeclareMathOperator{\ntfAC}{\mathcal{A}^{\mathcal{C}}}
\DeclareMathOperator{\ntfC}{\mathcal{C}}
\DeclareMathOperator{\ntfAvgA}{\bar{\mathcal{A}}}
\presetkeys{todonotes}{inline}{} %

\usepackage{tcolorbox}
\tcbset{size=minimal}
\colorlet{AddedContent}{green!30}
\definecolor{dgreen}{rgb}{0.0, 0.0, 0.0}

\begin{document}

\title{Leveraging Self-Supervised Vision Transformers for {\color{dgreen}Segmentation-based} Transfer Function Design}

\author{Dominik Engel, Leon Sick and Timo Ropinski
\thanks{Authors are with Visual Computing, Ulm University:~~\href{https://viscom.uni-ulm.de}{viscom.uni-ulm.de}}
\thanks{Corresponding author email:~~\href{mailto:research@dominikengel.com}{research@dominikengel.com}}%
\thanks{Manuscript received August 12, 2023; revised April 23, 2024.}}

\markboth{Accepted at IEEE Transactions on Visualization and Computer Graphics 2024  DOI: 10.1109/TVCG.2024.3401755}%
{Engel \MakeLowercase{\textit{et al.}}: Leveraging Self-Supervised Vision Transformers for Segmentation-based Transfer Function Design}

\IEEEpubid{}

\maketitle

\begin{abstract}
In volume rendering, transfer functions are used to classify structures of interest, and to assign optical properties such as color and opacity. They are commonly defined as 1D or 2D functions that map simple features to these optical properties. As the process of designing a transfer function is typically tedious and unintuitive, several approaches have been proposed for their interactive specification. In this paper, we present a novel method to define transfer functions for volume rendering by leveraging the feature extraction capabilities of self-supervised pre-trained vision transformers. To design a transfer function, users simply select the structures of interest in a slice viewer, and our method automatically selects similar structures based on the high-level features extracted by the neural network. Contrary to previous learning-based transfer function approaches, our method does not require training of models and allows for quick inference, enabling an interactive exploration of the volume data. Our approach reduces the amount of necessary annotations by interactively informing the user about the current classification, so they can focus on annotating the structures of interest that still require annotation. In practice, this allows users to design transfer functions within seconds, instead of minutes. We compare our method to existing learning-based approaches in terms of annotation and compute time, as well as with respect to segmentation accuracy. Our \textbf{\href{https://youtu.be/kTPBCYJtEJc}{accompanying video}} showcases the interactivity and effectiveness of our method.
\end{abstract}


\section{Introduction}
\IEEEPARstart{V}{isualizing} volumetric scientific data relies on a mapping of the underlying data to optical properties. In volume rendering, we call this mapping a \emph{transfer function}~(TF)~\cite{ljung2016state}. On scalar data, the simplest way to define a TF is by directly mapping the intensity of the input modality to optical properties, such as color and opacity. While such 1D TFs are simple to define and modify, they are inherently local and fail to extract semantically coherent regions that do not share a specific voxel value. Similarly, such simple TFs fail to separate different structures that share a value range. 

A plethora of work improves on this by extending the input space of the TF to 2D, including gradient magnitude~\cite{kniss2002multidimensional} or other possibly more complex local features~\cite{hladuvka2000curvature, haidacher2010volume,lundstrom2005extending}, usually at the cost of increasing the complexity of the TF definition and the user interface.
Another line of work proposes the collection of \emph{annotations} within slices, before training classifiers on the collected examples to predict which structures the remaining voxels belong to~\cite{tzeng2005intelligent, soundararajan2015learning, cheng2018deep}. Such an approach keeps the TF definition and user interface simple, but typically comes at the cost of losing interactivity, as these approaches require fitting of the annotated data points and inference for the remaining volume, which is prohibitively slow for existing approaches~\cite{tzeng2005intelligent, soundararajan2015learning, cheng2018deep}. As a result, these approaches feel more like a three-step process with an annotation phase, fitting \& inference phase, and a viewing phase.

In this work, we adopt the annotation-driven TF design paradigm, but enable an interactive process that gives immediate feedback upon user annotations. To achieve this, we leverage the features of a self-supervised Vision Transformer~(ViT) to identify structures matching the users annotations. 
Such networks are trained on millions of images with the goal of learning meaningful representations for all kind of different structures seen in those images. The sheer scale of the data and compute used in these pre-trainings leads to networks that produce meaningful features for all kinds of inputs~\cite{caron2021emerging}, including scientific data like CT or MRI. As a result these ViTs have been shown to perform very well in object discovery~\cite{wang2022self, hamilton2022unsupervised} and generally learn representations that are easily discriminated~\cite{caron2021emerging}. Using the semantically relevant features from the ViT, we identify the remaining voxels of a structure using feature similarity to compute a similarity map $\ntfS$. This approach is fast and can even run on CPU while maintaining interactivity. 

To utilize these self-supervised pre-trained ViTs in the 3D domain brings several challenges that we address in our paper. First, these networks are trained on 2D data, so we need a strategy to extract meaningful features from 3D volumetric data. Second, as a result of the input patching in ViTs, the features we extract are of comparatively low resolution that prohibit high visual quality when rendered directly. We address those issues by extracting features slice-wise along multiple axes, before merging the resulting 2D features to a 3D feature volume. To combat the low resolution we propose a refinement step that increases the resolution of our similarity maps and adapts to the underlying intensity volume. To achieve this we propose a 3D extension to the Fast Bilateral Solver~\cite{barron2016fast}.

In summary our method enables the following workflow: We start with a short pre-processing stage ($\approx 1-3$ minutes) to extract the feature maps. After feature extraction our method is interactive and allows users to explore the volume structures through annotation. Once a structure of interest is fully discovered, users can enable the refinement step ($\approx 0.5$ second) to increase the resolution and visual quality in the 3D rendering.

\IEEEpubidadjcol
\vspace{4mm}

To achieve this, we make the following contributions:
\begin{itemize}
    \item We propose a simple and fast, yet effective solution to leverage only neural network features to select and visualize volume structures from very few annotations.
    \item We enable an interactive annotation-guided transfer function design process with instant feedback after each annotation.
    \item To extract robust and discriminative features from volume data that serve as a basis for our annotation process, we leverage a frozen self-supervised Vision Transformer. We further propose a merging scheme to combine the extracted 2D feature maps into a 3D feature volume.
    \item We introduce a 3D extension to the Fast Bilateral Solver~\cite{barron2016fast} for refinement of our annotated similarity volumes.
\end{itemize}
We make the source code to our approach publicly available.\footnote{\url{https://dominikengel.com/vit-tf}}

\section{Related Work}
\subsection{Transfer function design} 
There has been a lot of work on designing transfer functions using different features, from simple 1D transfer functions based on intensity~\cite{drebin1988volume}, over 2D TFs based on gradients~\cite{kniss2002multidimensional} or segmentation maps~\cite{hadwiger2003high, bruckner2007style}. For example, Hladuvka\etal~\cite{hladuvka2000curvature} propose the use of curvature-based TFs, which is later built upon by Kindlmann\etal~\cite{kindlmann2003curvature} and Hadwiger\etal~\cite{hadwiger2005real}. Other works incorporate statistics about a voxel's local neighborhood~\cite{haidacher2010volume} or local frequency distribution~\cite{lundstrom2005extending, lundstrom2006local, lundstrom2006alpha}. Another line of work uses dimensionality reduction to utilize high-dimensional features in common 1D or 2D widgets~\cite{kniss2005statistically, haidacher2008information, kim2010dimensionality}. An extensive overview of these methods can be found in the survey by Ljung\etal~\cite{ljung2016state}.

\subsection{Learning-assisted transfer functions} 
The line of work on transfer functions most related to our approach deals with approaches that employ machine learning methods during the design process. Tzeng\etal~\cite{tzeng2005intelligent} pioneered the idea of collecting annotations from the users to offload the classification to a machine learning model. In their work they propose to first let users annotate slices of raw data, before training simple models like small neural networks and support vector machines (SVM) to classify the acquired data. In a similar fashion, Soundararajan and Schultz~\cite{soundararajan2015learning} provide a comparison of different classifiers for such a framework. Specifically they compared Gaussian Naive Bayes, k Nearest Neighbor, SVMs, neural nets and Random Forests (RF), where they found Random Forests to perform best. As features to their model they combine voxel intensity, intensity of neighboring voxels, gradient magnitude and voxel position to a feature vector of length $11$, for each voxel.

Zhou and Hansen~\cite{zhou2013transfer} propose probing of volume data using slice annotations to automatically generate 2D transfer functions using kernel density estimation. They use dimensionality reductions to project multivariate data and let users control the transfer function through a 2D Gaussian widget and a parallel coordinates plot. In a later work~\cite{zhou2014guideme}, they further introduce selection using a lasso tool to probe the slice views. 

De moura Pinto and Freitas~\cite{de2007design} propose the first unsupervised method, Kohonen Maps, to reduce the dimensionality of the high-dimensional TF space to enable TF design through common widgets.

Later, Cheng\etal~\cite{cheng2018deep} proposed to train convolutional neural networks (CNN) to extract high-level features. The CNN is trained for voxel-wise classification, and its predictions are used as input to marching cubes to generate a geometry. The extracted features are further ordered, so that users could define TFs based on characteristic features in a 1D TF widget. Their approach, however, requires labeled volumes to train the CNN, which drastically increases the computational cost.

Hong\etal~\cite{hong2019dnn} train a generative adversarial network~\cite{goodfellow2014generative} to predict rendered views from a view point, a rendering from this viewpoint that uses a trivial density to opacity mapping, and a goal image that conveys the style of the rendering (i.e. the mapping aspect of the TF). This approach however needs to be trained very costly for each volume and can barely be considered interactive even when deployed on their 8-GPU multiprocessing node.

Compared to this prior work, our approach brings several advantages. In contrast to the proposed supervised approaches that require large amounts of labeled training data, we leverage the generalized feature extraction capabilities of self-supervised pre-trained models and require no further training. This saves both the time needed for extensive annotation and training time, while enabling off-the-shelf application on a wide range of domains.
The annotation requirements in our approach are lightweight in comparison, since the only annotations we need are collected during the interactive transfer function design process, where the user clicks on the structures they would like to see in the rendering. Contrary to the annotation process of the other methods, our annotations are instantly followed up with feedback showing which structures were selected, eliminating the guess work for the amount of necessary annotations and the waiting time to evaluate the resulting selection.

\subsection{Self-supervised pre-training} 
Recently, several methods have made progress towards enabling the pre-training of vision models with unlabeled data~\cite{bao2021beit, caron2020unsupervised, caron2021emerging, chen2020improved, chen2020simple, chen2021empirical, he2020momentum, he2022masked,mitrovic2020representation,tomasev2022pushing,xie2022simmim,zhou2021ibot}. Chen\etal~\cite{chen2020simple} introduce an effective augmentation strategy to create multiple alternating versions of an image that are consequently fed through an encoder network and a projection head. Using this output, they compute a contrastive loss that learns to map images containing the same object closer together in the latent space. To tackle the problem of batch-size dependency for approaches of this kind, Caron\etal~\cite{caron2020unsupervised} propose an intermediate clustering of the latent representations by computing image codes and assigning them to cluster prototypes using the Sinkhorn-Knopp~\cite{cuturi2013sinkhorn} algorithm. Following the proposal of Vision Transformers~\cite{dosovitskiy2020image}, Caron\etal~\cite{caron2021emerging} have introduced DINO, a self-supervised model trained with a student-teacher knowledge distillation process. In their publication, they discover that ViTs can learn semantically-relevant structures in their intermediate features when pre-trained on unlabeled data with their method. In Section~\ref{sec:method}, we detail how we exploit this property to propose our ViT-based transfer function. Contrary to contrastive approaches, Bao\etal~\cite{bao2021beit} and He\etal~\cite{he2022masked} paved the way for self-supervised vision pre-training with masked-image-modeling approaches. In general, their approaches mask a portion of the input patches to the ViT and try to predict the masked patches and reconstruct the full input image, resulting in learned representations highly effective for model fine-tuning on several relevant tasks. Most recently, Assran\etal~\cite{assran2023self} have proposed an image-based joint-embedding predictive architecture (I-JEPA). Their approach provides the model with a context block, from which it is tasked to predict several target blocks in a single image. The learned representations have proven to be especially valuable for linear evaluations.

\color{dgreen}
\subsection{Segmentation methods}
The problem of segmentation has been tackled with a variety of approaches.%
Various works have proposed approaches to segment natural 2D images by annotating points in an interactive fashion~\cite{chen2022focalclick, wei2023focusedclick, li2023interactiveclick}. Li\etal~\cite{li2023interactiveclick} introduce a cross-modal vision transformer that takes as input the natural image and click annotations and employs cross-attention to learn from both modalities. In contrast to their method, our approach does not require a model training. Recently, ViTs have also successfully been applied to the problem of 2D medical image segmentation~\cite{liu2023optimizing, du2023mdvit, huang2021missformer, li2023lvit, li2023transformer}. Liu\etal~\cite{liu2023optimizing} modify a Swin UNet and add convolutional operations to preserve spatial locality. Du\etal~\cite{du2023mdvit} %
train a ViT on multiple domains using domain adapters and incorporate mutual knowledge distillation across domains. Huang\etal~\cite{huang2021missformer} introduce MISSFormer, for which they %
use an enhanced transformer context bridge and an enhanced transformer block to better capture long-range dependencies and local context. Furthermore, Li\etal~\cite{li2023lvit} propose a vision-language approach to medical image segmentation by combining image features and features from BERT-embedded medical text captions. %

For 3D medical image segmentation,  Hatamizadeh\etal~\cite{hatamizadeh2022unetr} proposed UNETR, a 3D transformer-based UNet. Their approach uses a transformer encoder on the 3D patches, followed by a decoder that uses convolution operations. Hatamizadeh\etal~\cite{hatamizadeh2021swin} also propose a hierarchical counterpart based on the popular Swin Transformer~\cite{liu2021swin}. Beyer\etal~\cite{beyer2022survey} compare a variety of interactive approaches that require training after annotation collection. Work by Liu\etal~\cite{liu2022isegformer} introduces iSegFormer, an interactive segmentation transformer for 3D knee MR images where the user inputs clicks and iteratively refines the prediction with more annotations. Their model is trained on both image and click embeddings, and in a class-agnostic fashion. Contrary to this, our approach requires neither training nor the embedding of click annotations. Also, their approach segments 2D slices, before relying on video segmentation propagation approaches to achieve 3D segmentation. This two step approach requires the propagation method to solve complex topologies based on just segmentation maps, whereas our approach merges features in 3D and avoids such propagation problems. Furthermore, due to the use of this propagation method, an inference of the full volume takes multiple seconds.

Recently, multiple works have built upon the Segment Anything (SAM)~\cite{kirillov2023segment} model to enable 3D medical segmentation~\cite{wang2023sam, chen2023ma, gong20233dsam}. One notable approach of these is SAM-Med3D~\cite{wang2023sam}. Wang\etal~modify the original SAM to have a 3D encoder, decoder and prompt encoder. Further, they perform a costly training data processing step to accumulate a large dataset, on which they train their supervised model. Our approach in contrast does not require any training, and therefore also no training data processing is needed. Further, since their approach is trained in a supervised fashion, it is at a higher risk of under-performing on unseen domains. Our feature encoder is pre-trained unsupervised, and hence does not suffer from this with similar severity. Further, work by Gong\etal \cite{gong20233dsam} have proposed a parameter-efficient adapters to enable SAM to accept 3D point prompts and decode the segmentation into a 3D volume. Also, their method requires further training of the proposed adapters.

\color{black}

\begin{figure*}[thb]
    \centering
    \includegraphics[width=\textwidth]{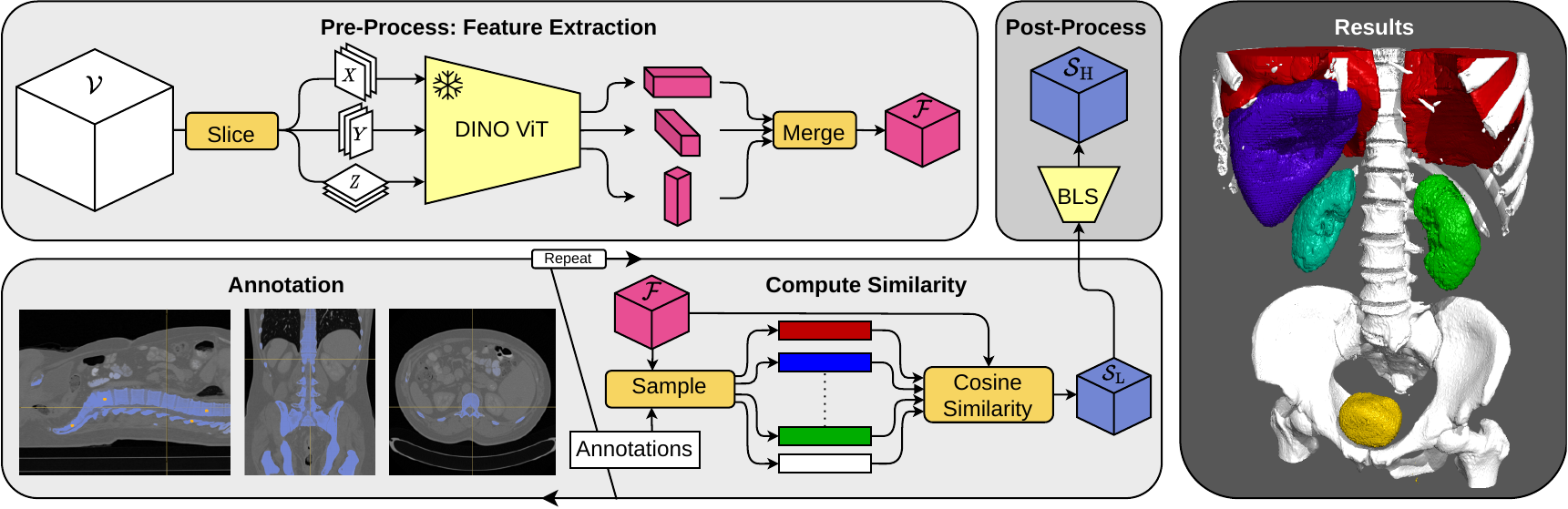}
    \caption{\textbf{Method Overview.} In the \textbf{Feature Extraction Pre-Processing} step, the volume data $\ntfV$ is \emph{sliced} along each axis and fed separately through the pre-trained DINO network. The resulting features are \emph{merged} into a feature volume $\ntfF$. Then, the user starts with \textbf{Annotation} in a slice viewer. Whenever the user annotates new voxels, we immediately \textbf{Compute Similarity} (blue highlights) of the annotated \emph{samples} (orange circles) with the feature volume $\ntfF$ (see Fig.~\ref{fig:gui} for a step-by-step visualization). 
    With the immediate feedback, the user can focus on the few regions that are missing after the initial annotations. Once the user is satisfied with $\ntfSL$, they can enable the \emph{bilateral solver (BLS)} as a \textbf{Post-Process} to obtain $\ntfSH$ with increased resolution. The whole process typically takes less than one minute in practice and is repeated for each class. Please watch the \textbf{\href{https://youtu.be/kTPBCYJtEJc}{supplemental video}} for a demonstration.}
    \label{fig:overview}
    \vspace{-3mm}
\end{figure*}

\section{Method}
\label{sec:method}
An overview of our approach is illustrated in Figure~\ref{fig:overview}. As a first step, our method extracts a feature volume $\ntfF$ using the pre-trained DINO~ViT~\cite{caron2021emerging} during pre-processing. This takes around one to two minutes on a consumer GPU and only needs to be performed once for a given volume $\ntfV$. During transfer function design, this feature volume $\ntfF$ is sampled at the locations that the user annotates. The sampled feature vectors are then compared to the full feature volume using \emph{cosine similarity} to obtain a similarity volume $\ntfSL$. When the user is satisfied with $\ntfSL$, it can be further refined using our 3D bilateral solver to obtain a high resolution similarity volume $\ntfSH$. The following subsections explain each of these steps, as well as the rendering procedure and user interface, in detail.

\subsection{Feature Extraction}\label{sec:feature-extraction}
Typically, transfer function design uses low-level and local features, like raw intensity, gradient magnitudes or local histograms. While these local features can be helpful in the separation of region of interest, they lack semantic meaning and may fail to capture the entirety of a region, putting the burden on the user through difficult interaction.
To combat this locality of the features, we propose the use of ViTs that by design relate different locations in the input to each other in their feature extraction. Specifically, we make use of self-supervised pre-trained ViTs. 

In our method, we use the DINO~\cite{caron2021emerging} ViT to extract representations. This network is originally trained on the RGB image domain. In order to feed our volumetric data through this 2D network, we first slice the volume along its three principal axes, then we replicate the slices to RGB and input them separately to DINO to extract representations. The resulting 2D representations are then again merged to form the 3D feature volume $\ntfF$. In the following, we first detail exactly what features we retrieve from the network, before describing the 2D to 3D process.

Specifically, we make use of the attention mechanism in the DINO ViT. Within the self-attention layers of the ViT, the feature maps from the previous block are fed through three linear layers, producing the \emph{key} ($K$), \emph{query} ($Q$) and \emph{value} ($V$) maps. In the attention mechanism, the $K$ and $Q$ are used to compute the attention matrix $A$ that determines the influence of the values $V$ for a specific attention head, that is finally passed on to the next layer:
$$ A = \text{softmax}(Q K^T / \sqrt{d}) $$
where $d$ is the feature dimension of the $Q,K,V$ maps divided by the number of heads in the attention layer.
{\color{dgreen} In our method we save the keys $K$ of the last self-attention layer in the ViT as feature map, as they represent semantic features that are designed to be matched to queries, which is exactly what we intend to do. 
This intuition is also supported by related work in unsupervised learning~\cite{wang2022self}.
In initial experiments, the $Q$ and $V$ feature maps performed very similar.}

In order to obtain the \emph{feature volume} $\ntfF$, we slice the input volume $\ntfV \in \mathbb{R}^{W\times H\times D}$ along each principal axis and feed the slices separately through the ViT network. {\color{dgreen}The resulting feature maps each have their un-sliced dimensions reduced by the patch size $p$ of the ViT, while keeping the sliced dimension unchanged}, resulting in:
\begin{align*}
  \ntfF_X & \in \mathbb{R}^{W \hspace{.5mm} \times \hspace{.5mm} H/p \hspace{.5mm} \times \hspace{.5mm} D/p \hspace{.5mm} \times \hspace{.5mm} F}, \\
  \ntfF_Y & \in \mathbb{R}^{W/p \hspace{.5mm} \times \hspace{.5mm} H \hspace{.5mm} \times \hspace{.5mm} D/p \hspace{.5mm} \times \hspace{.5mm} F}, \\
  \ntfF_Z & \in \mathbb{R}^{W/p \hspace{.5mm} \times \hspace{.5mm} H/p \hspace{.5mm} \times \hspace{.5mm} D \hspace{.5mm} \times \hspace{.5mm} F}
\end{align*}
In the following we call those reduced dimensions $W/p = W', H/p = H'$ and $D/p = D'$.
Having extracted the three stacks of feature maps, we need to merge them to one feature volume $\ntfF$. To obtain the merged $\ntfF$, these three features are first average pooled to the target dimensions and then averaged, resulting in a final resolution of $\ntfF \in \mathbb{R}^{W'\times H'\times D'\times F}$ with $F$ being the feature dimension, determined by the attention layers of the vision transformer. %

Since the feature maps have their spatial resolutions reduced by the patch size of the ViT, the resulting feature resolution may be quite low, depending on the input size. To enable control over the final dimensions $W', H', D'$, we optionally up-sample the images before we feed them to the ViT. This lets us choose arbitrary feature dimensions, but is restricted by the available GPU memory, as larger inputs to the ViT result in higher memory usage. In practice, we resize input images to around $640\times 640$, resulting in feature maps with a spatial dimension of $80$, which has proven to be a sufficient granularity for many structures (compare Section~\ref{sec:featres}, Appendix).

In our approach, we use the DINO~\cite{caron2021emerging} ViT-S/8 network, which has a patch size of $p=8$ and produces a $F=384$ - dimensional feature vector for each voxel in the feature grid. {\color{dgreen} We choose this network as it has been shown to extract meaningful features from many domains, while not being specifically trained for one.} It fits on a consumer GPU (RTX 2070, 8GB VRAM) and we can typically extract feature volumes of the size $\ntfF~\in~\mathbb{R}^{80\times 80\times 80\times 384}$. Larger transformer models like a ViT-B or ViT-L quickly require a prohibitive amount of GPU memory. They also typically come with an even larger patch size, thus decreasing the spatial resolution of the feature maps significantly. Similarly, newer models like the DINOv2~\cite{oquab2023dinov2} only come with a larger patch sizes and are therefore not considered for practical reasons.

\subsection{Computing Similarity Maps}\label{sec:similarity-computation}
After the feature volume $\ntfF$ is extracted and the user has made a first annotation (more details on the annotation interface in Section~\ref{sec:annotation-interface}), we compute how similar the annotated voxel is to each feature voxel in $\ntfF$. Intuitively, this can be thought of as querying the feature volume using singular features, closely matching the attention mechanism used during training of the network. 
Given a set of annotations $\ntfAC~\in~\mathbb{R}^{N\times 3}$ for class $\ntfC$, we compute the similarity as:
\begin{equation}
    \ntfSL^{\ntfC} = \max \left( \frac{1}{|\ntfAC|}\sum_{a \in \ntfAC} \sum_{\ntfF_i \in \ntfF}\frac{\ntfF_a \cdot \ntfF_i}{\left\| \ntfF_a \right\|_2 \left\| \ntfF_i \right\|_2}, 0 \right) \label{eq:cosine-similarity}
\end{equation}
where the resulting similarity $\ntfSL^{\ntfC} \in [0,1]^{W'\times H'\times D'}$ has the same spatial dimensions as $\ntfF$. This similarity computation is lightweight and only takes a few milliseconds on either CPU or GPU. This allows for immediate feedback to the user, thus we show an updated $\ntfSL$ right after an annotation is placed, enabling an interactive annotation process, where the user can make informed decisions about where to place further annotations.

Depending on the structure of interest, our similarity map may detect multiple occurrences of a structure withing a volume, i.e. two kidneys in a human CT, even when only one of them is annotated. This behavior follows directly from the global nature of the attention-based features. This aspect is especially useful to explore similar structures within a volume, however it often forbids the selection of just a single occurrence. 
\color{dgreen}
To combat this, the user can optionally use a \emph{proximity} parameter $\tilde{p} \in [0,1]$, which scales the similarity map $\ntfSL$ with $\mathcal{P}$, based on the distance to the closest annotation, allowing to select more spatially local structures if desired (see kidneys in Figure~\ref{fig:overview}):
$$ \mathcal{P}(x) = \max_{a \in \ntfAC} e^{-10 \tilde{p} |x - a|} \text{, for locations } x \in \mathbb{R}^3$$
\color{black}

\subsection{Post-Processing Similarity Maps}\label{sec:bilateral-solver}
As the initially computed low resolution similarity maps $\ntfSL$ lack the voxel-precise details required for a high visual fidelity when rendering, we propose a post-processing refinement step to  1) up-sample the similarity map and 2) adapt it to the surfaces seen in raw intensities in $\ntfV$. 
To achieve this, we implement a 3D version of the Fast Bilateral Solver (BLS)~\cite{barron2016fast}. The BLS is an edge-aware smoothing technique, similar to a bilateral filter, that considers a separate reference image to determine the degree of smoothing. We extend the approach to 3D by adding a z-component to each vertex in the bilateral grid. We use the 3D BLS to adapt our predicted similarity map to the edges of the underlying raw volume. 
Specifically, we first up-sample $\ntfSL$ tri-linearly to match the resolution of $\ntfV$, then we crop the regions where $\ntfS > \tau$ to discard low-similarity regions, before solving for a smoothed $\ntfSH$ using the according region from $\ntfV$ as reference for edge-awareness. As a threshold for cropping, we empirically choose $\tau = 0.25$.

Note that the spatial resolution of $\ntfSH$ can be chosen anywhere between the resolution of $\ntfF$ and $\ntfV$, enabling a trade-off between resolution/quality and speed. We typically choose the resolution of $\ntfSH$ at $256^3$ or $512^3$, depending on the class and the actual size of the structure, as this determines the size of the crop and therefore the running time. {\color{dgreen} Our current implementation of the solver runs on CPU and takes around $0.4$ seconds to process a $256^3$ volume and 
around $5.3$ seconds to process an $512^3$ volume on an Intel i7-8700K}. Since this post-processing is only run once after all annotations are placed, we can maintain an interactive experience. {\color{dgreen}The effect of this post-processing can be seen in the right two columns of Figure~\ref{fig:gui}, Figure~\ref{fig:comp-featres} and in the Appendix.}

\subsection{Rendering of Similarity Maps}\label{sec:rendering}
In order to visualize the volumetric data, we perform iso-surface raycasting on the similarity volumes $\ntfS$. During the interactive annotation, we only display $\ntfSL$, which can then be switched to $\ntfSH$ after post-processing when the annotation process is complete.
The raycasting approach steps through the volume until the similarity is above the iso-value defined for the according class $\ntfC$. Once the similarity increases over the iso-value, we perform a binary search to find the exact intersection of the ray and the iso-surface. After the surface is found, we blend its color onto the output buffer using forward compositing, before continuing with the raycasting until an early ray termination threshold is reached. Each point on the surface is shaded using the Phong shading model, together with a shadow ray cast towards the light source.

\begin{figure*}[thb]
    \centering
    \includegraphics[width=\textwidth]{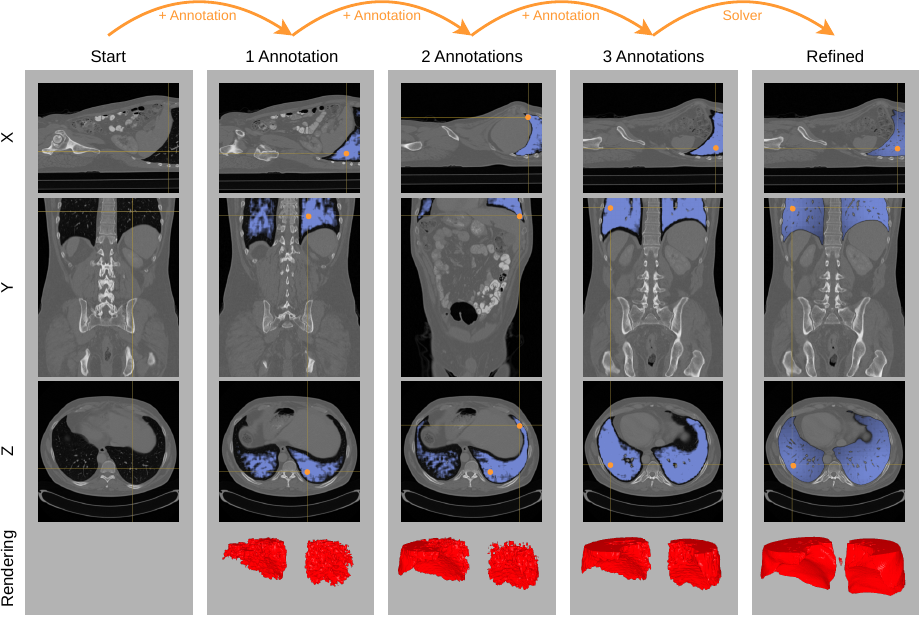}
    \caption{\textbf{Annotation Interface.} The user is presented with a slice viewer and a 3D rendering. Annotations can be either brushed using the mouse or set using individual points. After an annotation is set, the similarity map $\ntfSL$ is computed and displayed (blue) together with the annotation positions (orange circles). The 3D view displays an iso-surface rendering of $\ntfSL$. The similarity map informs the user where further annotations are required to fully segment the desired region. After just 3 annotations, the lung is mostly detected, and we can refine this result using the bilateral solver to obtain $\ntfSH$.}
    \label{fig:gui}
    \vspace{-3mm}
\end{figure*}

\subsection{Annotation Interface}\label{sec:annotation-interface}
The Annotation Interface is shown in Figure~\ref{fig:gui} and consists of a slice viewer for the three axes, as well as a canvas displaying the 3D rendering. The user can set annotations within the slice views, either by brushing lines or selecting individual points. After each annotation, all views are immediately updated, showing where previous annotations were set (orange points), as well as the current similarity map $\ntfSL$ to indicate which regions are already well recognized. This allows the user to make an informed decision about where to put further annotations, enabling users to quickly mark all regions of interest with just a few annotations, typically less than 10 per class, resulting in a fast TF design process. Misplaced annotations can be removed using a delete brush.

In addition to the slice viewer and 3D rendering, the user has an interface that allows adding and removing classes. For each class, the user can select a color and opacity used for rendering, as well two parameters. The first is the \emph{iso-value} slider threshold the similarity map. This effectively controls how \emph{semantically similar} voxels must be to the annotations. Further, users have a \emph{proximity} slider to restrict the predicted similarity to be \emph{spatially close} to the annotations. Lastly there is a checkbox to enable the 3D bilateral solver, i.e. the post-processing. With the bilateral solver come several parameters that are optionally configurable, namely $\sigma_\text{spatial}, \sigma_\text{chroma}, \sigma_\text{luma}$ from the original approach, which rarely need adjustment and are typically hidden in our GUI. {\color{dgreen} The full interface can be seen in Figure~\ref{fig:full-gui} and our \textbf{\href{https://youtu.be/kTPBCYJtEJc}{accompanying video}}.}

\begin{figure*}[thb]
    \centering
    \begin{subfigure}{0.26\textwidth}
        \includegraphics[width=\textwidth]{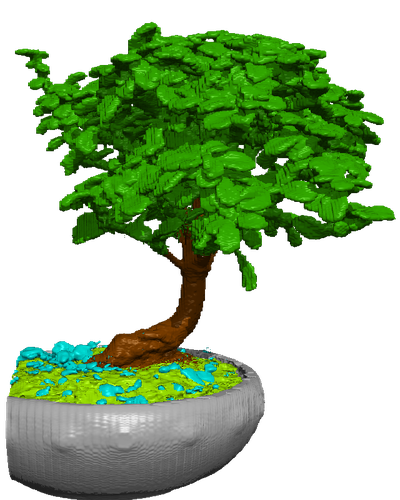}
    \end{subfigure}%
    \hfill
    \begin{subfigure}{0.20\textwidth}
    \includegraphics[width=\textwidth]{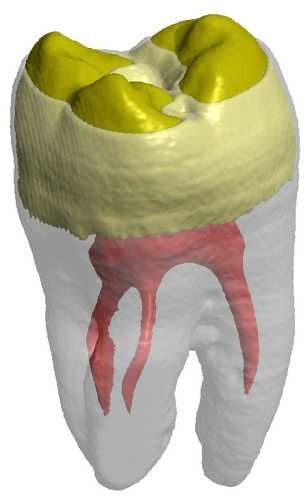}
    \end{subfigure}%
    \hfill
    \begin{subfigure}{0.49\textwidth}
    \includegraphics[width=\textwidth]{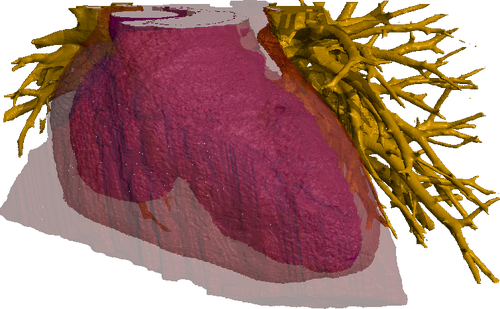}
    \end{subfigure}%
    \caption{\textbf{Qualitative Results.} We apply our method to various volume datasets, namely \textsc{Bonsai}, \textsc{Tooth} and the \textsc{MRI Heart}. Each of the classes required between 3 and 9 annotations.}
    \label{fig:various}
    \vspace{-3mm}
\end{figure*}

\section{Experiments}\label{sec:experiments}
In the following subsections, we perform several experiments to evaluate our approach. First, we look at  qualitative results, where we show renderings of different datasets and modalities, as well as a visual comparison to related work. Then we present a quantitative evaluation based on the \textsc{CT-ORG}~\cite{rister2020ctorg} segmentation dataset, where we also compare our approach to related work. 
In those experiments, we show how our approach compares to other methods, even when using three orders of magnitude fewer annotations. We further investigate the relevance of the resolution of the extracted feature volume $\ntfF$, 
{\color{dgreen} and lastly we perform a user study to assess the usability of our presented method.}

For the comparisons, we re-implemented the best performing approaches by Soundararajan and Schultz~\cite{soundararajan2015learning}, specifically their support vector machine (SVM) and random forests (RF). We chose this work for comparison, because it is reproducible due to their use of the classifiers by scikit-learn~\cite{pedregosa2011scikit}. It is also the most related to our approach, as they actively collect annotations from slice views, similar to our approach. Note that since their approach relies on direct classification of voxels, it requires a background class. When using our interactively collected annotations in their approach, we additionally draw samples at random from the background, matching the number of annotations of our most annotated class.
\color{dgreen}
As a second comparison, we use SAM-Med3D~\cite{wang2023sam}, the state-of-the-art click-to-segment SAM-derivative for 3D medical data.

Lastly we present additional experiments on (1) maintaining topological consistency by using our 2D $\rightarrow$ 3D merging strategy, (2) combining DINO features with SVMs and RFs, and (3) a quality assessment of how our refinement step restores fine details that are lost in low resolution similarity maps $\ntfSL$, in the Appendix.
\color{black}

\begin{figure}[!thb]
    \vspace{-2mm}
    \centering
    \begin{subfigure}{0.4\linewidth}
        \includegraphics[width=\textwidth]{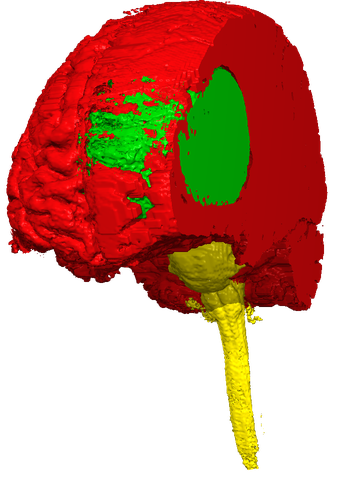}
    \end{subfigure}%
    \begin{subfigure}{0.48\linewidth}
        \includegraphics[width=\textwidth]{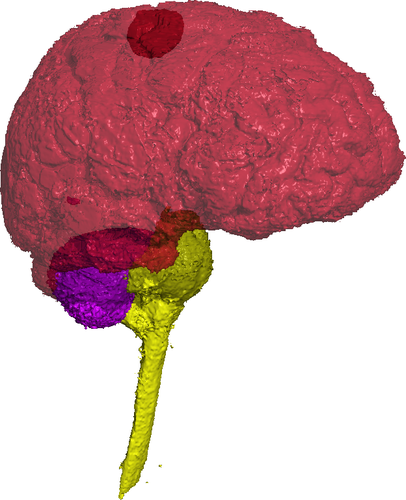}
    \end{subfigure}%
    \caption{\textbf{Qualitative Results on MRI} (\textsc{VisContest2010})}
    \label{fig:mri}
    \vspace{-3mm}
\end{figure}

\begin{figure}[!thb]
    \centering
    \includegraphics[width=\linewidth]{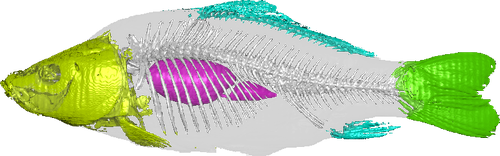}
    \caption{\textbf{Qualitative Results on CT} (\textsc{Carp})}
    \label{fig:carp}
    \vspace{-3mm}
\end{figure}

\subsection{Visual Results on Different Modalities}
In this experiment, we apply our method to various datasets to show its applicability on different types of data. Figure~\ref{fig:various} shows renderings of three different datasets. For the \textsc{Bonsai} and \textsc{MRI Heart} datasets we use on average 5 annotations per class. The \textsc{Tooth} required 6 annotations for the pulpa, 9 for the enamel and 8 for the dentin.
Figure~\ref{fig:mri} shows results on the \textsc{VisContest2010} dataset, specifically the case 2 T1 MRI pre surgery, where we require 17 annotations for the brain matter, 8 for the tumor and 5 for the brain stem and the post-surgery case 2 T1 where about 5 annotations per class suffice.
We also apply our approach on animal scans, as shown in 
Figures~\ref{fig:carp}~and~\ref{fig:jarv}.

Results for the \textsc{Bonsai} and \textsc{Tooth} dataset are also reported by Soundararajan and Schultz~\cite{soundararajan2015learning}. Since they require thousands of annotations, we could not feasibly reproduce their exact results here for a direct comparison, however they can be viewed in their work. When using their approach with the few annotations we require, all their models fail to produce a meaningful result, as the surrounding air is falsely predicted to belong to one of the classes, occluding any structure of interest.

As can be seen in these figures our approach manages to define meaningful transfer functions from just very few annotations and works for a variety of structures and modalities.

\subsection{Visual Comparison to Soundararajan\etal~\cite{soundararajan2015learning}}
\begin{figure*}[thb]
    \begin{subfigure}{0.24\textwidth}
    \includegraphics[width=\textwidth]{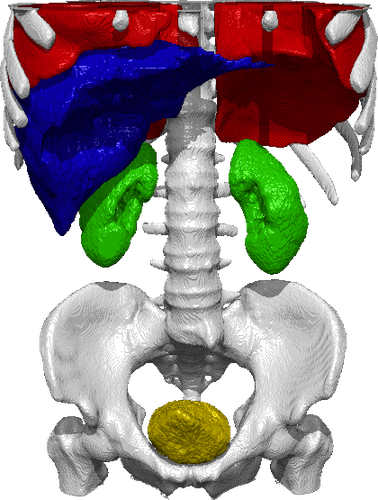}
    \caption{Ground Truth}
    \end{subfigure}%
    \begin{subfigure}{0.24\textwidth}
    \includegraphics[width=\textwidth]{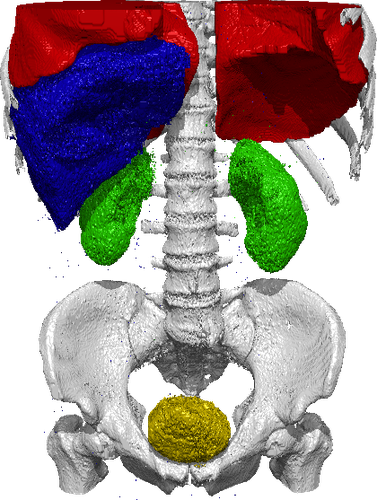}
    \caption{Ours $\ntfAvgA=5.2$}
    \end{subfigure}%
    \begin{subfigure}{0.26\textwidth}
    \includegraphics[width=\textwidth]{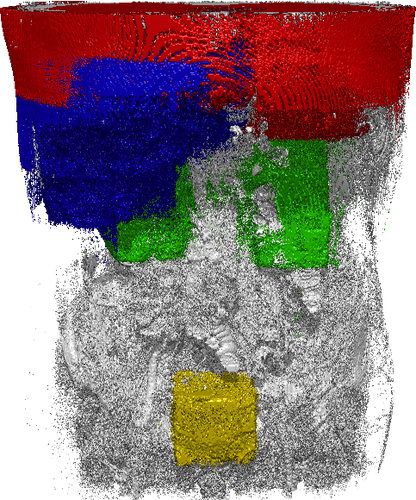}
    \caption{RF $\ntfAvgA = 8192$}
    \end{subfigure}%
    \begin{subfigure}{0.26\textwidth}
    \includegraphics[width=\textwidth]{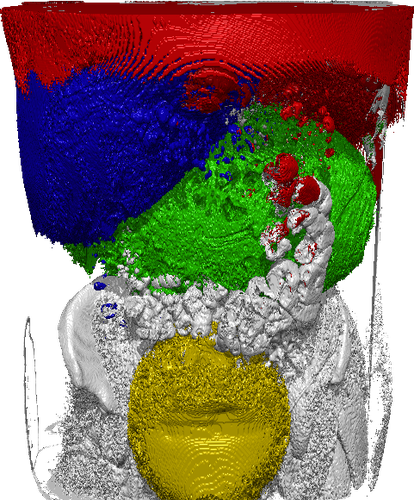}
    \caption{SVM $\ntfAvgA = 8192$}
    \end{subfigure}%
    \caption{\textbf{Visual Comparison} to the SVM and RF approach by Soundararajan\etal~\cite{soundararajan2015learning} on \textsc{CT-ORG}. This visualization matches the predictions in Table~\ref{tab:comp-samples} and shows the RF and SVM with 8192 training samples per class, while Ours only uses the interactively collected annotations (on average $\ntfAvgA=5.2$ annotations per class).}
    \label{fig:comparison-ctorg}
    \vspace{-2mm}
\end{figure*}

We compare our approach to the aforementioned SVM and RF approaches on the \textsc{CT-ORG}~\cite{rister2020ctorg} dataset. This dataset has high-resolution CT scans of human torsos, as well as ground truth segmentations for the liver, bladder, lung, kidney and bones. 
Figure~\ref{fig:comparison-ctorg} compares the ground truth segmentation to our approach using on average $\ntfAvgA = 5.2$ annotations per class, as well as results from Soundararajan\etal~\cite{soundararajan2015learning}. For their approach, we show the models trained with $8192$ samples per class, as this large amount of annotations produced the best results for their approach. When using just he $\ntfAvgA = 5.2$ annotations per class that we use for our approach, their methods fail to produce a meaningful result. In order to choose the annotations to train their approach, we randomly sample $8192$ annotations per class from the ground truth labels. In Figure~\ref{fig:comparison-ctorg} their methods use around $1500\times$ the amount of annotations compared to ours.

\subsection{Quantitative Comparisons}
We compare our method quantitatively to the SVM and RF approach by Soundararajan\etal\cite{soundararajan2015learning} and to SAM-Med3D~\cite{wang2023sam} on the \textsc{CT-ORG}~\cite{rister2020ctorg} dataset. This experiment reports segmentation metrics that match the visual results in Figure~\ref{fig:comparison-ctorg}. To compute such metrics, we need to convert our similarity maps $\ntfSH$ to classification decisions for each voxel. For this, we threshold the similarity maps for each class using the iso-value used for rendering, and in the case that a voxel would be assigned multiple classes, we choose the one with the highest similarity value.

Table~\ref{tab:comp-classes} shows results for the Precision, Recall, F1-Score and Intersection over Union (IoU) for the different classes using our set of interactively collected annotations.
Table~\ref{tab:comp-samples} further shows results for an increasing amount of samples for the SVM and RF approach. \textbf{Ours} in this table still only uses the $\ntfAvgA = 5.2$ annotations per class, and the table shows that our approach is superior to the classifier-based approach even when they receive an unreasonably large amount of annotations. 
Figure~\ref{fig:iou-plot} further shows how our approach performs in terms of mean IoU, compared to the increasing amount of annotations used to train the RF and SVM. 

\begin{table}[bht]
    \centering
    \begin{tabular}{llccccc}
    \toprule
     \hspace{3mm} & Method & Liver & Bladder & Lung & Kidney & Bone \\
    \midrule
    \multirow{5}{*}{\rotatebox{90}{Precision}} 
        & \textbf{Ours} &  0.977  & \textbf{0.999} & \textbf{0.988} & 0.997  & \textbf{0.985} \\
        & SAM-Med3D (t) & \textbf{0.988} & 0.993 & 0.986 & \textbf{0.999} & 0.927 \\
        & SAM-Med3D (o) & 0.974 & 0.992 & 0.850 & 0.971 & 0.872 \\
        & RF &  0.173  & 0.034 & 0.608 & 0.033  & 0.157 \\
        & SVM &  0.0  & 0.019 & 0.0 & 0.0  & 0.134 \\
     \midrule
     \multirow{5}{*}{\rotatebox{90}{Recall}}
        & \textbf{Ours} &  \textbf{0.978}  & \textbf{0.999} & \textbf{0.992} & 0.996  & \textbf{0.988} \\
        & SAM-Med3D (t) & \textbf{0.978} & 0.973 & 0.988 & \textbf{0.998} & 0.935 \\
        & SAM-Med3D (o) & 0.972 & 0.983 & 0.988 & 0.942 & 0.967 \\
        & RF &  0.046  & 0.953 & 0.378 & 0.124  & 0.781 \\
        & SVM &  0.0  & \textbf{0.999} & 0.0 & 0.0  & 0.707 \\
     \midrule
     \multirow{5}{*}{\rotatebox{90}{F1 Score}}
        & \textbf{Ours} &  0.978  & \textbf{0.999} & \textbf{0.990} & \textbf{0.997}  & \textbf{0.987} \\
        & SAM-Med3D (t) & \textbf{0.984} & 0.986 & 0.987 & \textbf{0.997} & 0.931 \\
        & SAM-Med3D (o) & 0.973 & 0.987 & 0.914 & 0.956 & 0.917 \\
        & RF &  0.073  & 0.066 & 0.466 & 0.053  & 0.262 \\
        & SVM &  0.0  & 0.038 & 0.0 & 0.0  & 0.225 \\
     \midrule
     \multirow{5}{*}{\rotatebox{90}{IoU}}
        & \textbf{Ours} &  0.957  & \textbf{0.999} & \textbf{0.980} & 0.994  & \textbf{0.975} \\
        & SAM-Med3D (t) & \textbf{0.968} & 0.972 & 0.975 & \textbf{0.997} & 0.872 \\
        & SAM-Med3D (o) & 0.948 & 0.975 & 0.847 & 0.917 & 0.847 \\
        & RF &  0.038  & 0.034 & 0.304 & 0.027  & 0.151 \\
        & SVM &  0.0  & 0.019 & 0.0 & 0.0  & 0.127 \\
     \bottomrule
    \end{tabular}\vspace{2mm}
    \caption{\color{dgreen}\textbf{Segmentation Metrics by class on \textsc{CT-ORG}.} We compare to Soundararajan\etal~\cite{soundararajan2015learning} and SAM-Med3D~\cite{wang2023sam} using the annotations gathered during interactive annotation. On average, each class has $\ntfAvgA = 5.2$ annotations.}
    \label{tab:comp-classes}
    \vspace{-3mm}
\end{table}

\begin{figure}[htb]
    \centering
    \includegraphics[width=\linewidth]{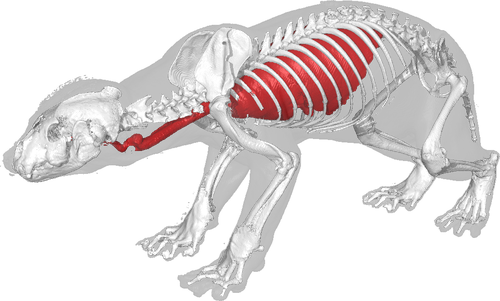}
    \caption{\textbf{Qualitative Results} for the \textsc{Järv} (wolverine) dataset.}
    \label{fig:jarv}
\end{figure}

\subsection{Impact of feature volume resolution}\label{sec:featres}
As described in Section~\ref{sec:feature-extraction}, we can control the resolution of the feature volumes $\ntfF$ that we extract from the ViT. By resizing the slices fed into the network, the resulting feature resolution can be increased at the cost of increased computational demand and memory footprint. Generally a higher resolution feature map allows for more granularity in the initial similarity maps $\ntfSL$, and could allow for better detection of fine structures. In order to understand the importance of the resolution of $\ntfSL$ we annotate the ribs in the \textsc{CT-ORG} dataset with 9 annotations and compute similarity maps from feature volumes of different resolutions. We then tune similarity thresholds individually, before applying the bilateral solver for refinement. Figure~\ref{fig:comp-featres} shows renderings of the resulting similarity maps for features of resolution $64^3$, $80^3$ and $96^3$ and their according refined similarities. {\color{dgreen} We further test our method's ability to detect very fine fish bones in the Appendix.}

\section{Discussion}\label{sec:discussion}
\subsection{Visual Results}
As shown in Figures~\ref{fig:various}-\ref{fig:jarv} our approach is able to design meaningful transfer functions using only a few annotations. Our method could separate different structures well and works on different kinds of data, like CT and MRI scans of very different objects. Some structures show small visual artifacts, caused by the iso-surface rendering of not fully completed structures. This occurs with insufficient $\ntfSL$, as described in the Appendix.

\subsection{Segmentation Performance}
In order to get a quantitative measure of our method's performance, we applied it on the \textsc{CT-ORG} dataset, which has segmentation ground truth that we can use to compute segmentation metrics. Table~\ref{tab:comp-classes} and~\ref{tab:comp-samples} show that our method is able to extract the five different types of organs with relatively few annotations. Overall the liver was the most difficult to segment, meaning it required the most tuning of iso-value and proximity parameters. {\color{dgreen}We compare our results to the state-of-the-art 3D medical SAM method, SAM-Med3D~\cite{wang2023sam}, whereby we compare to two different variants of this method. One trained specifically on \emph{organ} data, and the other being the more general \emph{turbo} variant (SAM-Med3D (o) and (t) in the tables). As can be seen, overall our method performs quite similar to SAM-Med3D and outperforms it just slightly in terms of segmentation quality.}

\begin{figure}[bh]
    \centering
    \includegraphics[width=\linewidth]{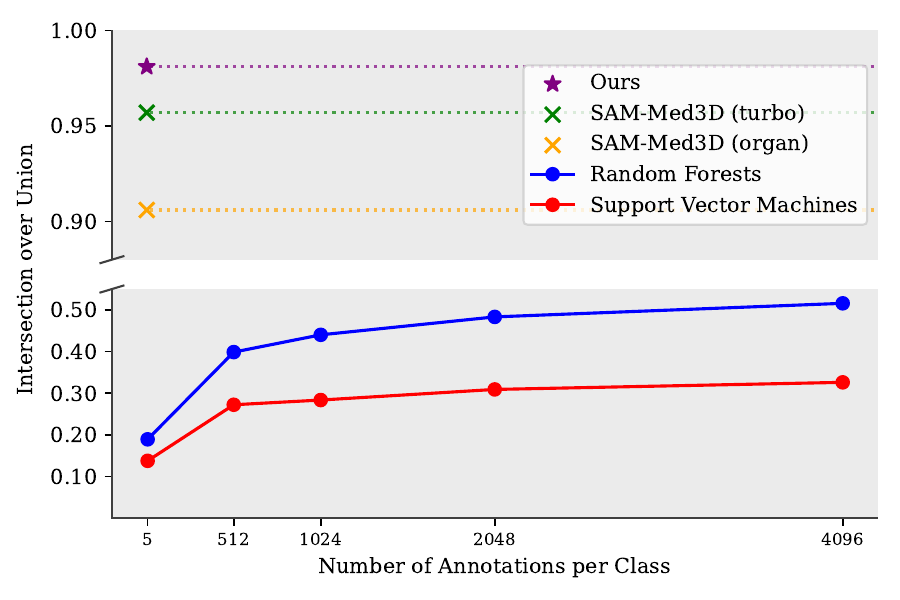}
    \caption{\color{dgreen}\textbf{Intersection over Union on \textsc{CT-ORG}}. We compare the IoU of our approach using the interactively collected annotations ($\ntfAvgA= 5.2$) with SAM-Med3D~\cite{wang2023sam} and the SVM and RF approach by Soundararajan\etal~\cite{soundararajan2015learning}. Our approach has superior IoU with just $5.2$ annotations per class on average, even compared to thousands of annotations for SVMs and RFs.}
    \label{fig:iou-plot}
\end{figure}

Compared to the SVM and RF proposed by Soundararajan\etal~\cite{soundararajan2015learning} we find our segmentation performance favorable, even when increasing the amount of annotations for the SVM and RF by three orders of magnitude. Figure~\ref{fig:iou-plot} shows that the SVM and RF approaches improve with an increased amount of annotations, although they plateau well below our mean IoU of $0.981$. The SVM and RF approach are also quite slow in comparison, as summarized in Table~\ref{tab:time-measurements}.

\begin{table}[t]
    \centering
    \begin{tabular}{llccccc}
    \toprule
     $\ntfAvgA$ &Method & Acc. & Prec. & Recall & F1 & mIoU \\
    \midrule
    \multirow{5}{*}{5.2}
     & \textbf{Ours}   &  \textbf{0.981}  & \textbf{0.989} & \textbf{0.990} & \textbf{0.990} & \textbf{0.981} \\
     & SAM-Med3D (t) & 0.964 & 0.980 & 0.979 & 0.977 & 0.957 \\
     & SAM-Med3D (o) & 0.914 & 0.932 & 0.976 & 0.950 & 0.906 \\
     & RF     &  0.722  & 0.329 & 0.509 & 0.296  & 0.218 \\
     & SVM    &  0.669  & 0.181 & 0.405 & 0.180  & 0.139 \\
     \midrule
     \multirow{2}{*}{1024}
     & RF     &  0.855  & 0.472 & 0.931 & 0.587  & 0.440 \\
     & SVM    &  0.724  & 0.340 & 0.859 & 0.399  & 0.283 \\
     \multirow{2}{*}{2048}
     & RF     &  0.870  & 0.512 & 0.943 & 0.630  & 0.483 \\
     & SVM    &  0.750  & 0.356 & 0.883 & 0.425  & 0.306 \\
     \multirow{2}{*}{8192}
     & RF    &  0.901  & 0.567 & 0.962 & 0.686  & 0.545 \\
     & SVM    &  0.796  & 0.390 & 0.909 & 0.473  & 0.348 \\
     \bottomrule
    \end{tabular}\vspace{2mm}
    \caption{\color{dgreen}\textbf{Segmentation Metrics by Annotation Amount.} $\ntfAvgA$ denotes the number of annotations per class. We compare our method on \textsc{CT-ORG} with $\ntfAvgA=5.2$ interactively collected annotations to the SVM and RF approach by Soundararajan\etal~\cite{soundararajan2015learning} using varying amounts of annotations.}
    \label{tab:comp-samples}
\end{table}

\begin{table}[thb]
    \centering
    \begin{tabular}{llccc}
        \toprule
         Resolution & Method &  Pre-Proc. & Train & Inference \\
         \midrule
         $64^3$ & \textbf{Ours} & 13.5s & \textbf{-} & \textbf{0.01s} \\
         \midrule
         \multirow{3}{*}{$128^3$} & \textbf{Ours} +BLS & 14.4s & - & \textbf{0.10s} \\
         & SAM-Med3D (t)~\cite{wang2023sam} & 3s & - & 3.9s \\
         & SAM-Med3D (o)~\cite{wang2023sam} & 3s & - & 3.9s \\
         \midrule
         $256^3$ & \textbf{Ours} +BLS & 25.7s & - & \textbf{0.37s} \\
         \midrule
         \multirow{5}{*}{$512^3$} & \textbf{Ours} +BLS & 132.1s & - & \textbf{5.33s} \\
         & RF~\cite{soundararajan2015learning} \hfill $\ntfAvgA=5.2$ \hphantom{0}& - & 0.03s& 172s\\
         & SVM~\cite{soundararajan2015learning} \hfill $\ntfAvgA=5.2$ \hphantom{0}& - & 0.001s & 323s \\
         & RF~\cite{soundararajan2015learning} \hfill $\ntfAvgA=4096$ & - & 2s & 432s\\
         & SVM~\cite{soundararajan2015learning} \hfill $\ntfAvgA=4096$ & - & 9s & 71500s \\
         \bottomrule
    \end{tabular}
    \caption{\color{dgreen}\textbf{Time Measurements.} Numbers reported per class on \textsc{CT-ORG}. \textbf{Ours} extracts features once in the beginning, but needs no training. During annotation the inference time of \textbf{Ours} applies regardless of resolution, and is followed by a post-processing (\textbf{Ours} + BLS) that varies with resolution.}
    \label{tab:time-measurements}
\end{table}

\begin{figure}
    \centering
    \begin{subfigure}{0.32\linewidth}
        \includegraphics[width=\textwidth]{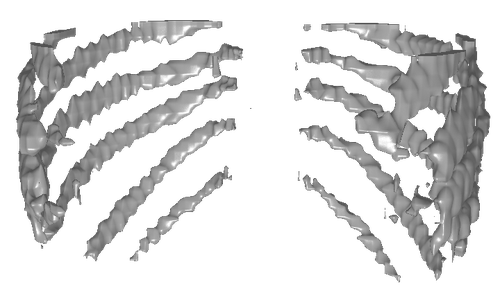}
        \includegraphics[width=\textwidth]{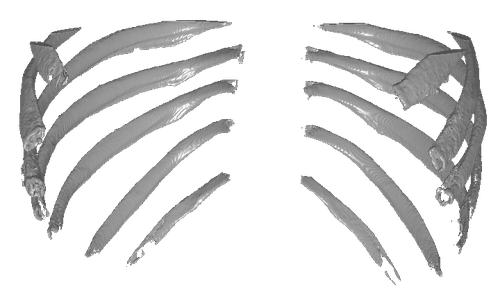}
        \caption{$64^3$}
    \end{subfigure}%
    \begin{subfigure}{0.32\linewidth}
        \includegraphics[width=\textwidth]{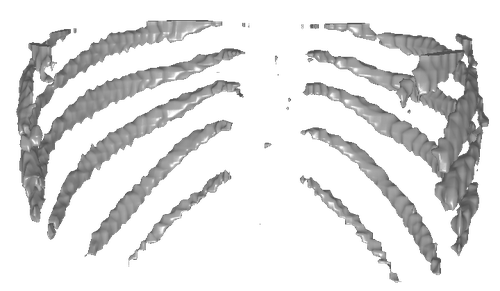}
        \includegraphics[width=\textwidth]{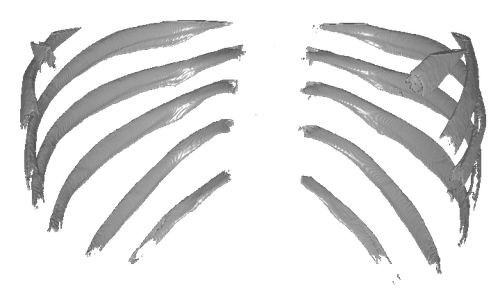}
        \caption{$80^3$}
    \end{subfigure}%
    \begin{subfigure}{0.32\linewidth}
        \includegraphics[width=\textwidth]{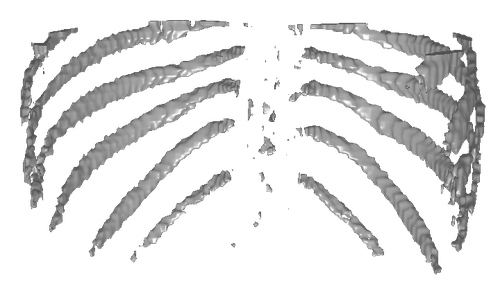}
        \includegraphics[width=\textwidth]{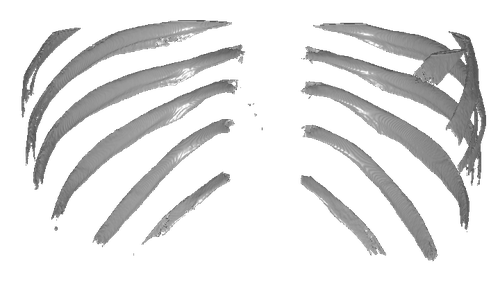}
        \caption{$96^3$}
    \end{subfigure}%
    \caption{\textbf{Comparison of Feature Resolutions.} Top row shows un-refined similarity maps at the given resolution, bottom row shows the results after refinement.}
    \label{fig:comp-featres}
\end{figure}

\subsection{Impact of feature volume resolution}
As shown in Figure~\ref{fig:comp-featres}, the resolution of $\ntfF$ has a visible impact on the un-refined similarity maps. We can see that higher feature resolutions provide less visual artifacts in the form of blockiness. However, all of the similarity maps managed to capture so much of the ribs, that the refinement step is able to completely select them in all cases, leaving the final refined results very similar. This makes clear that very high resolution feature maps are not necessary to obtain voxel-precise predictions. We found that as long as a structure is detected in $\ntfSL$, the refinement step can typically extract the structure of interest and is not very prone to the resolution of $\ntfSL$. In practice that enables our method to be useful on consumer GPUs, as 8GB of VRAM suffice to extract features of resolution $80^3$, whereas higher resolutions would quickly demand a prohibitive amount of VRAM to extract. 

\color{dgreen}
\subsection{User Study}
\begin{figure}
    \centering
    \includegraphics[width=\linewidth]{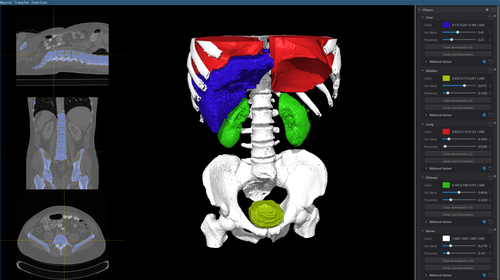}
    \caption{\color{dgreen}\textbf{Graphical User Interface.} The left side shows the slice annotation viewer depicted in Fig.~\ref{fig:gui}, center shows the 3D rendering and on the right users can define classes.}
    \label{fig:full-gui}
    \vspace{-3mm}
\end{figure}

In order to verify the usability of our approach, we performed a user-study with $N=12$ participants ($7$ male, $5$ female, average age $29.125$). Participants rated their familiarity with navigation of 3D software between $2$ and $4$ on a 5-point likert scale, with an average of $3.75$, however none of the participants was familiar with medical data or navigation using synchronized slice views. The participants were first briefly introduced to the user interface of our approach (compare Figure~\ref{fig:full-gui}), before being allowed to familiarize themselves with the controls (between 2-5 minutes). After this introduction participants were asked to segment the \emph{lung}, \emph{liver} and \emph{kidney} from the \textsc{CT-ORG} dataset. To solve this task, they were shown the ground truth segmentations beforehand, to ensure that they are able to identify the organs correctly within the CT scan. Lastly we asked the participants to rate our method using the System Usability Scale (SUS)~\cite{brooke1996sus}.

\begin{figure}
    \centering
    \includegraphics[width=\linewidth]{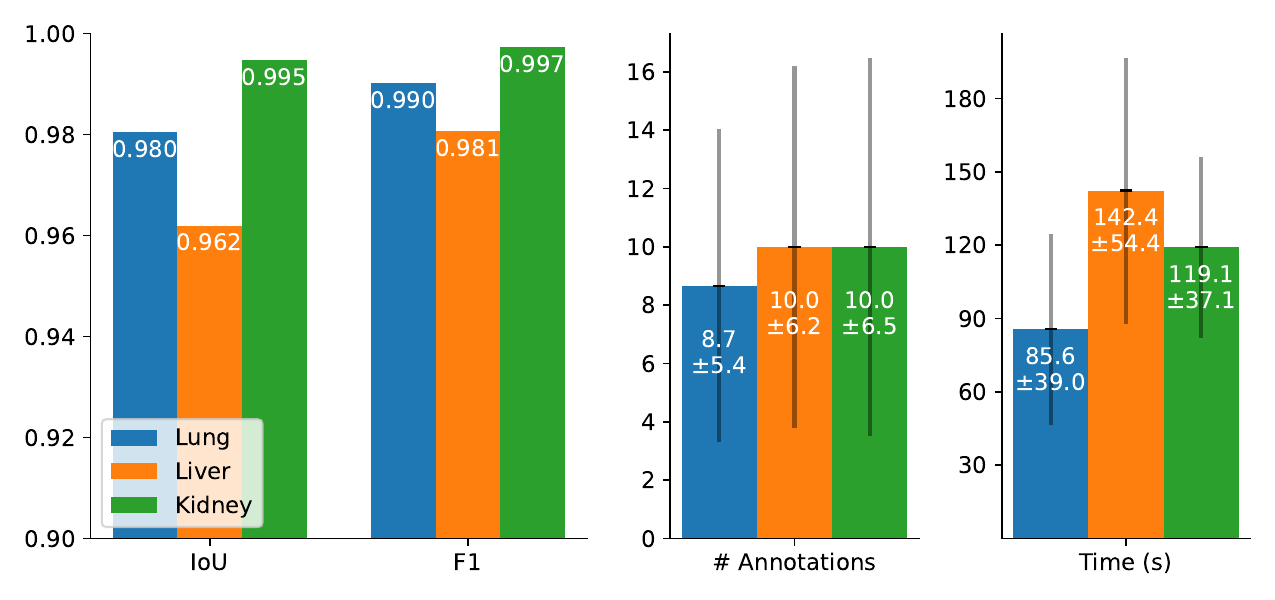}
    \caption{\color{dgreen}\textbf{User Study Results.} We report segmentation metrics compared to the ground truth, as well as number of click annotations and time. The error bars in the right plots indicate the standard deviation (omitted on the left, as it is $<$ 1e-3)}
    \label{fig:userstudy-metrics}
    \vspace{-3mm}
\end{figure}

The results of the segmentation task can be seen in Figure~\ref{fig:userstudy-metrics}. As it can be seen, the participants were able to achieve very strong segmentation results ($\text{IoU} > 0.95$) in about 1-3 minutes with on average 10 or less annotations. Note that participants were not asked to keep the number of annotations minimal, but were allowed to use our method as they see fit. All participants achieved very similar segmentation metrics (standard deviation $<$ 1e-3), indicating that all organs could be segmented precisely, regardless of the strategy (tuning iso-value / proximity vs. placing more annotations).

\begin{figure}
    \centering
    \includegraphics[width=\linewidth]{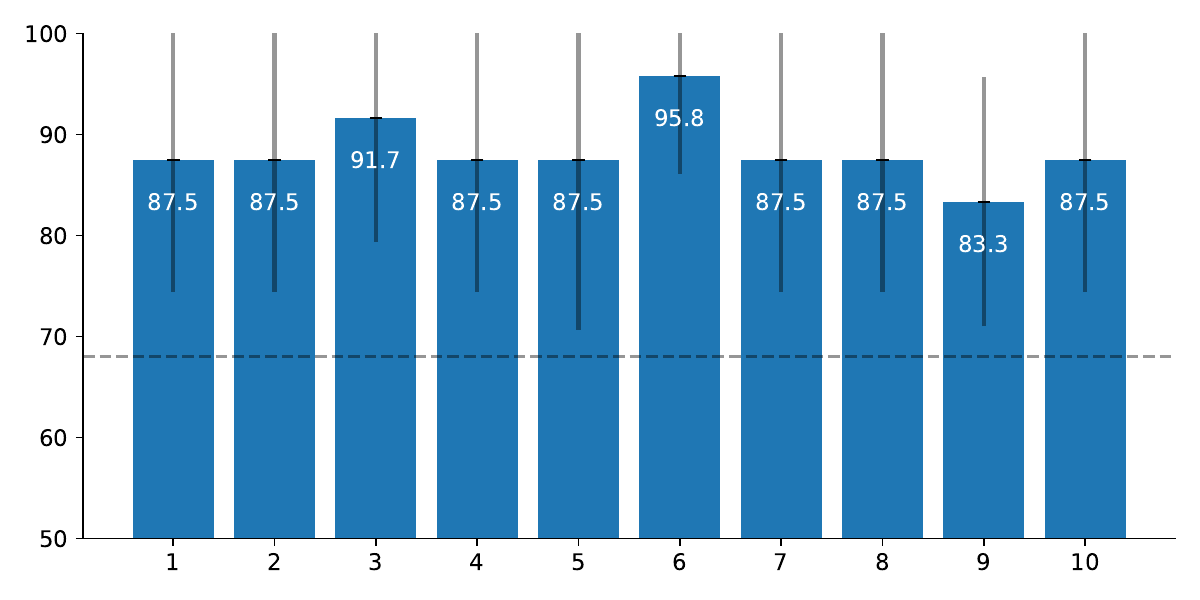}
    \caption{\color{dgreen}\textbf{Results of System Usability Scale}~\cite{brooke1996sus} per question.}
    \label{fig:sus}
    \vspace{-3mm}
\end{figure}

The results of the SUS questionnaire are displayed in Figure~\ref{fig:sus}. The overall SUS score is $88.25$ (of $100$), exceeding the average score of $68$~\cite{brooke2013sus}, indicating above average usability of our system. In the scope of the questionnaire we also gave participants the option to leave free-text comments on our approach. Generally our method was received well and is perceived "very helpful for medical segmentation", with a "clean UI with fast visual feedback, minimal extra sliders and high responsiveness". Nevertheless, one common suggestion for improvement that we received is the wish for "negative annotations", to be placed in regions to be excluded from a class. We agree with this suggestion and plan to tackle such a feature in future work.
\color{black}

\subsection{Limitations}
One limitation is that our pre-processing step, the feature extraction, can be quite memory intensive. Vision transformers require lots of memory, especially when we try to achieve high resolutions for $\ntfF$. To obtain a certain feature resolution, the input to the ViT must be scaled by the patch size. In practice this quickly exceeds the memory budget on consumer GPUs, as all the feature maps need to be saved for all three slicing directions and lastly be pooled to the desired feature size. While we have shown that our approach does not heavily rely on high resolutions of $\ntfSL$, this high memory requirement also prevents us currently from using larger transformer models, like the ViT-B or ViT-L or transformer models with higher patch sizes.

We further found that when selecting a structure within a volume, it may occur that our method recognizes more structures of similar appearance, that we may not want to select. An example for this is the bladder in the \textsc{CT-ORG} dataset. When annotated, other structures like the kidneys or surrounding tissue is often deemed similar, which is a common problem for many approaches, due to the similar intensities in a CT. While we can circumvent this to some extent by placing more annotations in the actual region of interest, this results in precisely choosing thresholds for the similarity map. We also implemented the option to use a connected components filter to discard disconnected components that are falsely detected to combat this problem, which works well for separated structures, like two kidneys (compare Figure~\ref{fig:overview}), but fails when the structures to be separated are too close to each other.

Lastly we find that when structures cannot be perfectly detected at their surfaces, the resulting renderings may show the block artifacts. We describe this problem in detail in the Appendix.

\subsection{Future Work}
In the future, we see several additions and improvements to an approach like ours. Firstly, the use of larger pre-trained transformers, as well as the option to retrieve higher resolution feature maps, would probably improve the method's performance significantly.

Another interesting direction to look into is using neural nets that are pre-trained to learn joint image and text embeddings, like CLIP~\cite{radford2021learning}, BLIP~\cite{li2022blip} or OpenCLIP~\cite{cherti2023reproducible}. Those networks are trained to produce similar features for images and matching text, and could enable our approach to use natural language queries to selected structures as part of the transfer function design process, in addition to spatial annotations.

\color{dgreen}
Lastly, we and several of the participants of our user study noted, that the notion of \emph{negative annotations} could prove useful to select structures of interest. There are several possibilities to implement such a mechanism and we plan to explore this idea in future work.
\color{black}

\section{Conclusion}
To conclude, we have presented a novel method for transfer function design, leveraging self-supervised pre-trained Vision Transformers. We show that the features of such a network can be used to design transfer functions by querying the feature map by singular feature vectors obtained through annotation. By giving the user immediate feedback on the obtained similarities for the current set of annotations, users can easily find regions that require further annotation to ultimately reduce the need for a large number of annotations. This enables users to create transfer functions for a structure of interest in seconds to minutes, and hence allows for quick visualization and exploration of volume datasets. In comparison to prior machine learning based transfer function approaches, our interface and annotation process is kept to a minimum, and we can avoid actually training a model, by just utilizing the features of the pre-trained network. Further, our method is quick enough to design transfer functions interactively, without requiring a separate annotation phase. To increase the visual quality of rendering our similarity maps, we propose a 3D extension to the fast bilateral solver~\cite{barron2016fast} that lets us up-sample similarity maps to a high resolution. Our approach can be easily extended in the future through the use of newer and larger networks, or even networks that produce features that can be queried by natural language.

\section*{Acknowledgments}
The annotation interface is implemented in the Inviwo~\cite{jonsson2019inviwo} framework, and renderings were produced using Inviwo.

\bibliographystyle{IEEEtran}
\bibliography{references}

\section*{Biography Section}
\vspace{-1cm}
\begin{IEEEbiography}[{\includegraphics[width=1in,height=1.25in,clip,keepaspectratio]{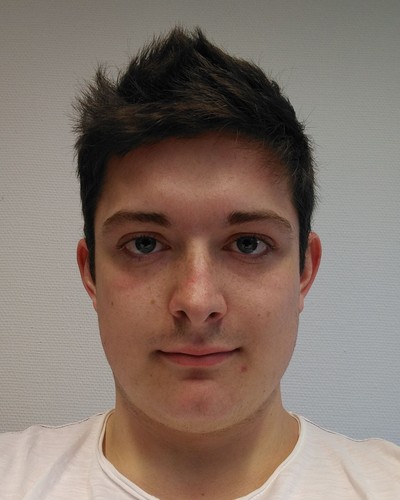}}]{Dominik~Engel}
is a Ph.D. student at Ulm University, Germany, where he previously received his B.Sc. and M.Sc. degrees in computer science. In 2018, he joined the Visual Computing research group. His research focuses on deep learning in visualization and computer graphics, differentiable and neural rendering.
\end{IEEEbiography}
\vspace{-12mm}
\begin{IEEEbiography}[{\includegraphics[width=1in,height=1.25in,clip,keepaspectratio]{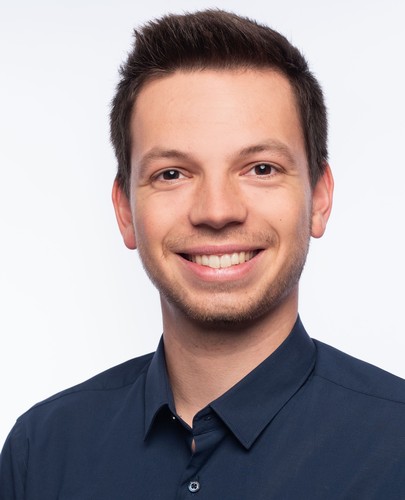}}]{Leon~Sick}
is a Ph.D. student at Ulm University and part of the Visual Computing Group. Before starting his Ph.D., he obtained his B.A. in International Business Administration from Aalen University of Applied Sciences and his M.Sc. in Business Information Technology from Konstanz University of Applied Sciences. His research is focused on self-supervised pre-training and unsupervised segmentation on 2D images.
\end{IEEEbiography}
\vspace{-12mm}
\begin{IEEEbiography}[{\includegraphics[width=1in,height=1.25in,clip,keepaspectratio]{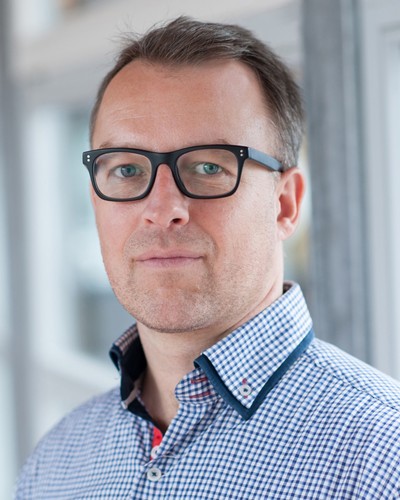}}]{Timo~Ropinski}
is a professor at Ulm University, heading the Visual Computing Group. Before moving to Ulm, he was Professor in Interactive Visualization at Linköping University, heading the Scientific Visualization Group. He received his Ph.D. in computer science in 2004 from the University of Münster, where he also completed his Habilitation in 2009. Currently, Timo serves as chair of the EG VCBM Steering Committee, and as an editorial board member of IEEE TVCG.
\end{IEEEbiography}

\vfill

\end{document}


\title{Leveraging Self-Supervised Vision Transformers for Segmentation-based Transfer Function Design (Appendix)}
\date{ }

\author{Dominik Engel\thanks{Corresponding author email:~~\href{mailto:research@dominikengel.com}{research@dominikengel.com}}%
, Leon Sick and Timo Ropinski
}

\maketitle

\section{BLS Input Requirements}\label{sec:discuss-refinement}
\begin{figure}[th]
    \centering
    \begin{subfigure}{0.48\linewidth}
        \begin{subfigure}{0.49\textwidth}
            \includegraphics[width=\textwidth]{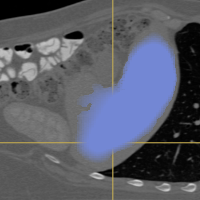}
        \end{subfigure}%
        \hfill
        \begin{subfigure}{0.49\textwidth}
            \includegraphics[width=\textwidth]{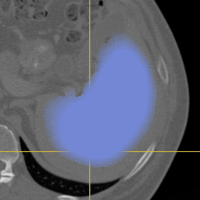}
        \end{subfigure}%
        \\
        \includegraphics[width=\textwidth]{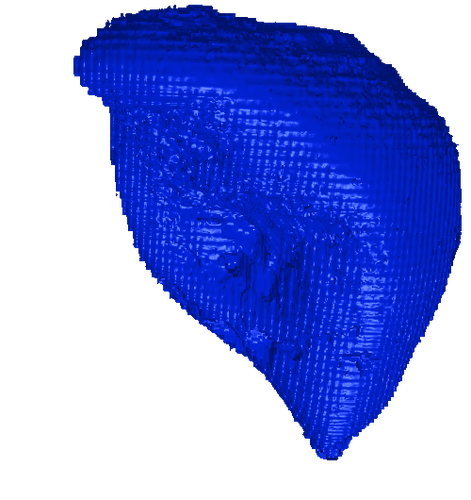}
        \caption{Insufficient $\ntfSL$}
    \end{subfigure}%
    \hfill
    \begin{subfigure}{0.48\linewidth}
        \begin{subfigure}{0.49\textwidth}
            \includegraphics[width=\textwidth]{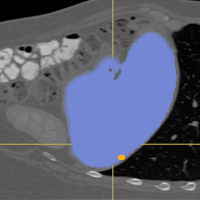}
        \end{subfigure}%
        \hfill
        \begin{subfigure}{0.49\textwidth}
            \includegraphics[width=\textwidth]{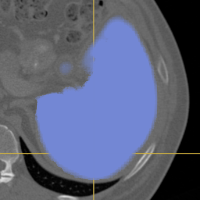}
        \end{subfigure}%
        \\
        \includegraphics[width=\textwidth]{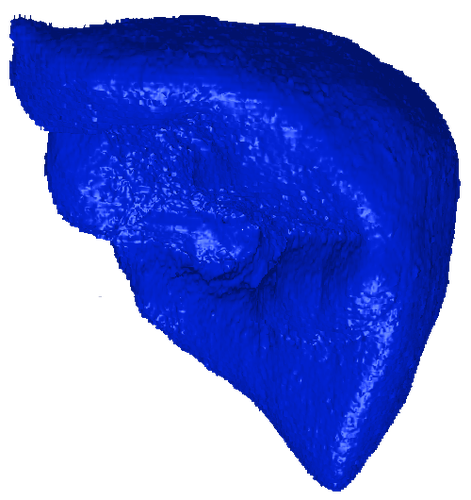}
        \caption{Sufficient $\ntfSL$}
    \end{subfigure}%
    \caption{\textbf{Refinement Artifacts with insufficient $\ntfSL$.} Our refinement step requires an input similarity map $\ntfSL$ with sufficient detection of the relevant structures to be effective, and results in block artifacts otherwise.}
    \label{fig:bls-artefacts}
    \vspace{3mm}
\end{figure}

To better understand the importance of the refinement step and its dependency on its input, the initial low resolution similarity $\ntfSL$, we produce an initial similarity map with insufficient annotation. This similarity map only captures the center region of the liver in the \textsc{CT-ORG}~\cite{rister2020ctorg} dataset, as shown in Figure~\ref{fig:bls-artefacts}. We now apply the refinement step on this incomplete similarity map to find out if the bilateral solver can complete the structure, and therefore if it can compensate for a lack of detection in $\ntfSL$.

In our testing we found that the refinement step is typically good at aligning the low resolution feature maps to the raw input, especially at the borders of the structure, while not being able to complete structures far beyond what is detected in $\ntfSL$. If a structure is not sufficiently detected in $\ntfSL$, the refinement step is unable to complete the structure. This is illustrated in Figure~\ref{fig:bls-artefacts}, where we refine a similarity map that has insufficient annotations and misses the borders of the liver. The refined similarity falls off smoothly towards the liver surface, resulting in block artifacts when rendered as an iso-surface. The block artifacts arise from the $\sigma$ parameters of the bilateral solver, that control the window used for blurring. We find that those block artifacts only occur, when the structure is not sufficiently detected in the low resolution similarity map. We conclude that the refinement step requires a sufficient detection of the full structure already in the low resolution similarity map, in order to produce surfaces without artifacts. This highlights the importance of both the initial similarities $\ntfSL$ and the refinement step.

\begin{figure*}[t]
    \centering
    \begin{subfigure}{0.2493\linewidth}
    \includegraphics[width=\textwidth]{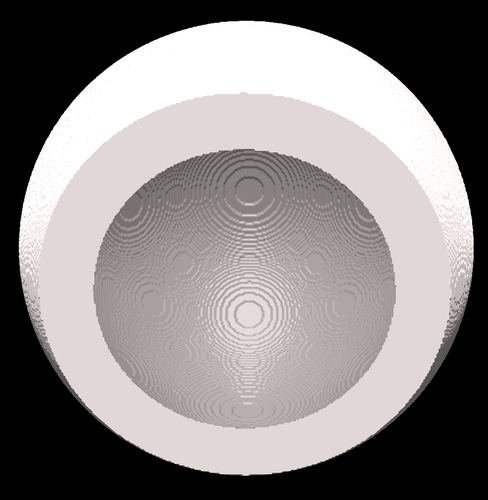}
    \end{subfigure}%
    \begin{subfigure}{0.2494\linewidth}
    \includegraphics[width=\textwidth]{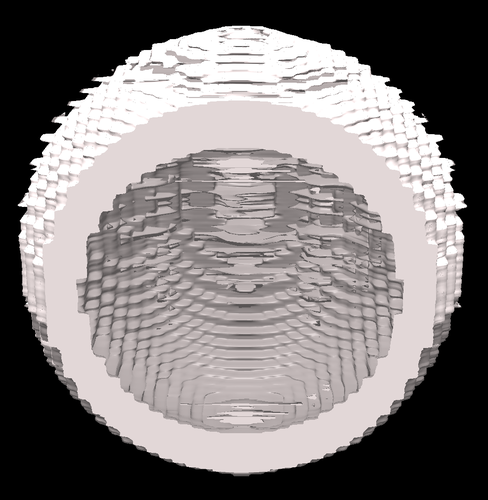}
    \end{subfigure}%
    \begin{subfigure}{0.25\linewidth}
    \includegraphics[width=\textwidth]{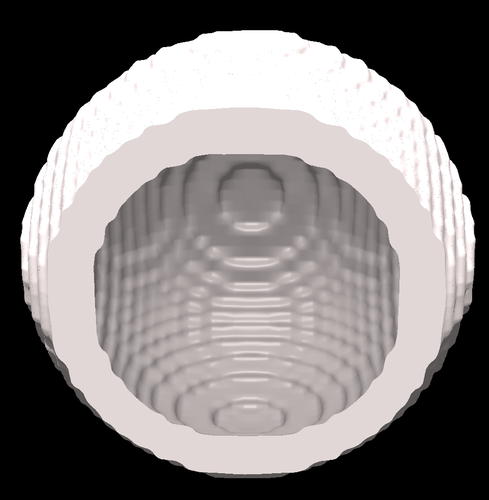}
    \end{subfigure}%
    \begin{subfigure}{0.2493\linewidth}
    \includegraphics[width=\textwidth]{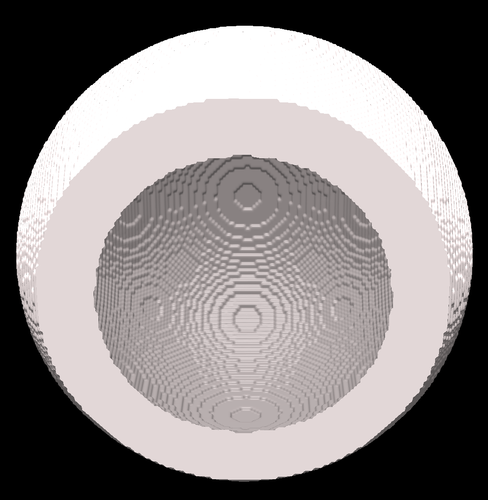}
    \end{subfigure}%
    \\
    \begin{subfigure}{0.25\linewidth}
    \includegraphics[width=\textwidth]{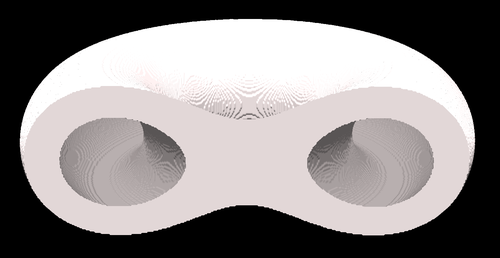}
    \caption{Ground Truth}
    \end{subfigure}%
    \begin{subfigure}{0.25\linewidth}
    \includegraphics[width=\textwidth]{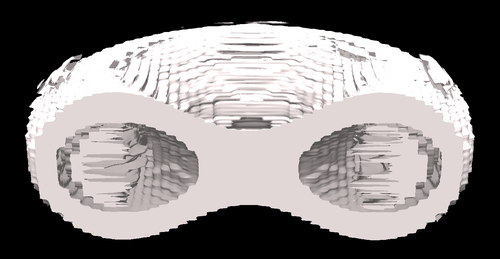}
    \caption{Z only}
    \label{fig:topo-z-only}
    \end{subfigure}%
    \begin{subfigure}{0.25\linewidth}
    \includegraphics[width=\textwidth]{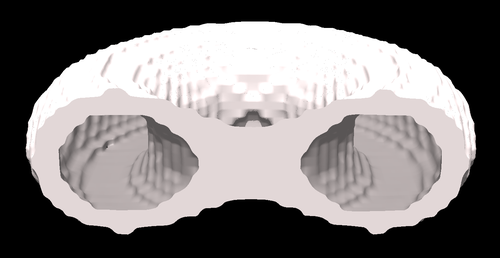}
    \caption{Ours}
    \label{fig:topo-xyz}
    \end{subfigure}%
    \begin{subfigure}{0.25\linewidth}
    \includegraphics[width=\textwidth]{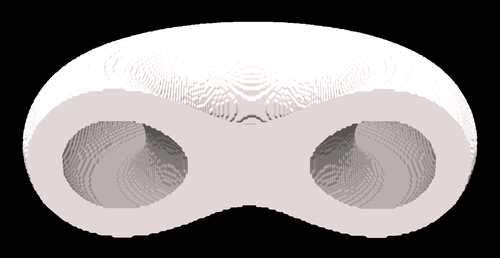}
    \caption{Ours + BLS}
    \end{subfigure}%
    \caption{\textbf{Topological Consistency.} Compare how simply using slice-wise features (b) results in topological inconsistencies, while our 2D~$\rightarrow$~3D strategy (c) provides all the necessary information to preserve the topology of the sphere and torus.}
    \label{fig:topo-consistency}
    \vspace{-3mm}
\end{figure*}

\section{Topological Consistency}
When extracting the features for our approach, we employ a 2D $\rightarrow$ 3D merging strategy to achieve per-voxel features that match similar structures throughout the whole volume. Many related approaches that employ such kind of 2D $\rightarrow$ 3D propagation suffer from topological inconsistencies. In this section, we investigate if such problems arise with our merging strategy. For our investigation we created two synthetic datasets, \textsc{Sphere} and \textsc{Torus} at the resolution $512^3$. We then compare how well our features propagate without our merging strategy, i.e. from just a single slicing direction, and with our merging strategy. The results of this experiment can be seen in Figure~\ref{fig:topo-consistency}. 

Fig.~\ref{fig:topo-z-only} shows that, when simply extracting features slice-wise, they suffice to recognize the full sphere \emph{with artifacts}, but introduce great errors inside the torus. Our merged features in Fig.~\ref{fig:topo-xyz} correctly predict the full 3D structure and avoid the artifacts produced by using slice-wise features. Note that the block artifacts visible in both the slice-wise and our merged approach are due to the spatial resolution of the feature map ($64^3$) and can be mostly removed using the Bilateral Solver (BLS).

We conclude that our approach to merging the X, Y and Z-sliced feature maps alleviates possible topological inconsistencies. Furthermore, since our implementation of the bilateral solver operates in 3D space, it is also able to resolve possible topological inconsistencies.

\section{DINO Features for SVM/RF}
Our approach makes use of very high-level global features extracted by the DINO~\cite{caron2021emerging} network. We argue that this feature map contains way more information compared to more local features, as utilized in e.g. the SVM or RF by Soundararajan~\cite{soundararajan2015learning}. We have therefore tried using the SVM and RF approach with this more global feature map as input.

The SVM approach could not make reasonable predictions based on the DINO feature map. With the background excluded, the SVM achieves an IoU of $0.24$ for the liver, $0.36$ for the bones and $0.0$ for the other classes in the \textsc{CT-ORG} dataset, resulting in a mean IoU of $0.09$. The SVM is also prohibitively slow to fit ($407$s) and infer ($2077$s) with the DINO features. Note that actual inference would take around $10\times$ as long and the IoUs would likely further decrease when including the background voxels.
The Random Forest is a lot faster to fit ($3$s) and infer ($2.5$s), however it also fails to achieve meaningful segmentation, even with drastically increased number of estimators. Neither the SVM, nor the RF surpass a training accuracy above chance.

\section{Number of Annotations}
For completeness we document the number of annotations required for each class in our demonstrated Figures of the main paper in Table~\ref{tab:num-annotations}.

\begin{table}[!htb]
    \centering
    \resizebox{\linewidth}{!}{
    \begin{tabular}{clc}
    \toprule
        Dataset & Class & \# Annotations \\
    \midrule
    \multirow{5}{*}{\shortstack{\textsc{Bonsai}\\ (Fig. 3)}} 
            & Pot & 4 \\
            & Stem & 9 \\
            & Soil & 3 \\
            & Stones & 3 \\
            & Leaves & 7 \\
    \midrule
    \multirow{3}{*}{\shortstack{\textsc{Tooth}\\ (Fig. 3)}}
            & Pulpa & 6 \\
            & Enamel & 9 \\
            & Dentin & 8 \\
    \midrule
    \multirow{3}{*}{\shortstack{\textsc{MRI Heart}\\ (Fig. 3)}}
            & Chambers & 7 \\
            & Septum & 3 \\
            & Arteries & 6 \\
    \midrule
    \multirow{3}{*}{\shortstack{MRI Brain (pre)\\ \textsc{VisContest2010}\\ (Fig. 4)}}
            & Brain Matter & 17 \\
            & Tumor & 8 \\
            & Stem & 5 \\
    \midrule
    \multirow{4}{*}{\shortstack{MRI Brain (post)\\ \textsc{VisContest2010}\\ (Fig. 4)}}
            & Brain Matter & 2 \\
            & Tumor & 6 \\
            & Stem & 1 \\
            & Cerebellum & 5 \\
    \midrule
    \multirow{6}{*}{\shortstack{\textsc{Carp}\\ (Fig. 5)}}
            & Fish Bones & 5 \\
            & Fish Bladder & 2 \\
            & Head & 5 \\
            & Top Fin & 5 \\
            & Bottom Fin & 2 \\
            & Tail Fin & 9 \\
    \midrule
    \multirow{5}{*}{\shortstack{\textsc{CT-ORG}\\ (Fig. 6)\\ (see Appendix Fig.~\ref{fig:ctorg-with-annotations})}} 
             & Liver & 6 \\
             & Bladder & 2 \\
             & Lung & 2 \\
             & Kidneys & 3 \\
             & Bones & 13 \\
    \midrule
    \multirow{4}{*}{\shortstack{\textsc{Järv}\\ (Fig. 7)}}
            & Bones & 24 \\
            & Lung & 6 \\
            & Trachea & 5 \\
            & Skin & 8 \\
    \bottomrule
    \end{tabular}}
    \caption{\textbf{Number of annotations per class} used for the Figures of the \textbf{main paper}.}
    \label{tab:num-annotations}
\end{table}

\begin{figure}
    \centering
    \includegraphics[width=\linewidth]{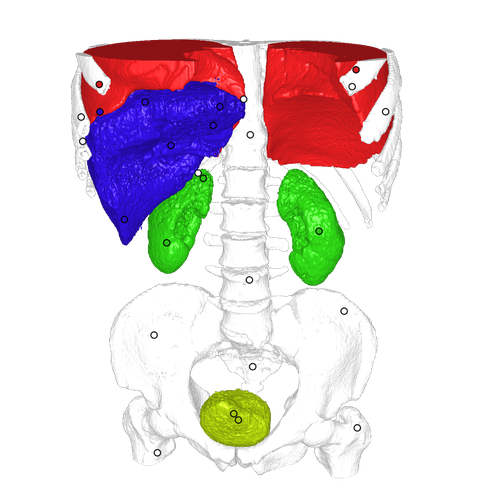}
    \caption{\textbf{CT-ORG with annotations visualized.}}
    \label{fig:ctorg-with-annotations}
\end{figure}

\section{Fine Details}
\begin{figure*}
    \begin{subfigure}{0.241\textwidth}
        \includegraphics[width=\linewidth]{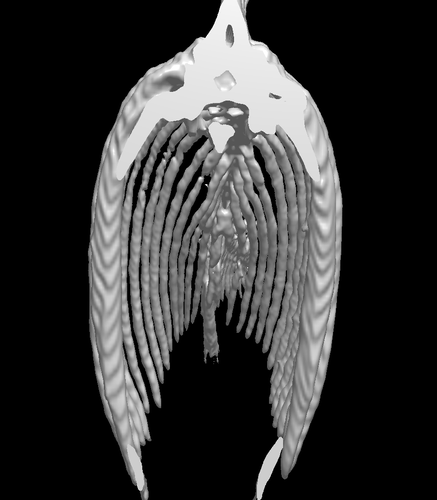}
    \end{subfigure}%
    \begin{subfigure}{0.482\textwidth}
        \includegraphics[width=\linewidth]{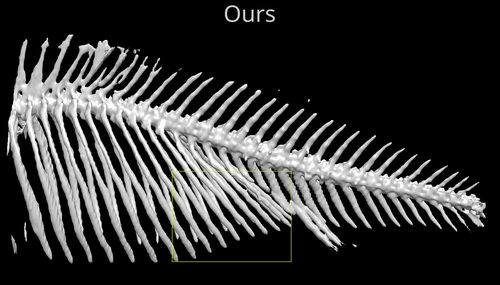}
    \end{subfigure}%
    \begin{subfigure}{0.275\textwidth}
        \includegraphics[width=\linewidth]{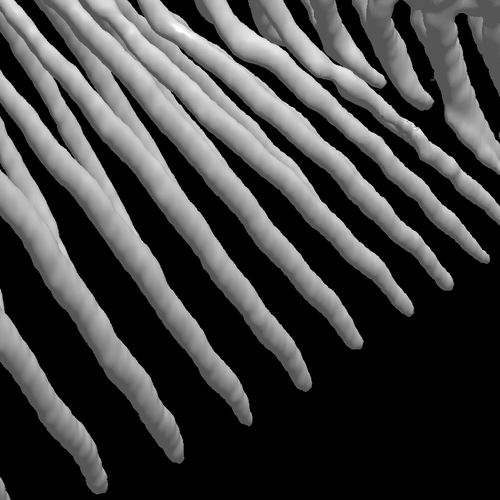}
    \end{subfigure}%
    \\
    \begin{subfigure}{0.241\textwidth}
        \includegraphics[width=\linewidth]{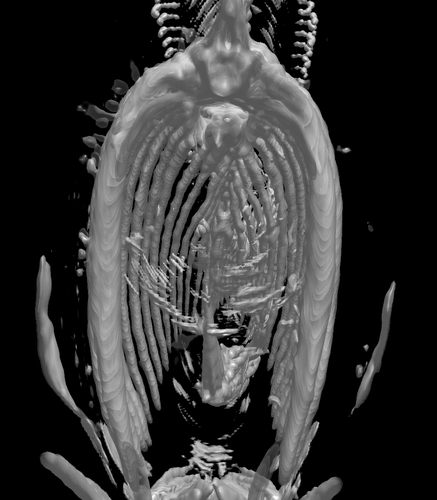}
    \end{subfigure}%
    \begin{subfigure}{0.482\textwidth}
        \includegraphics[width=\linewidth]{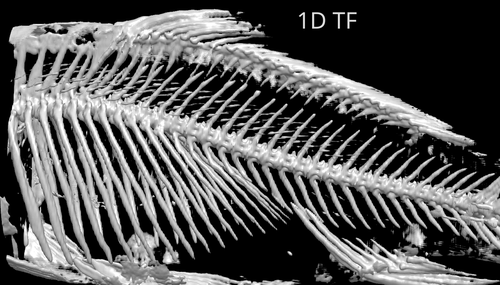}
    \end{subfigure}%
    \begin{subfigure}{0.275\textwidth}
        \includegraphics[width=\linewidth]{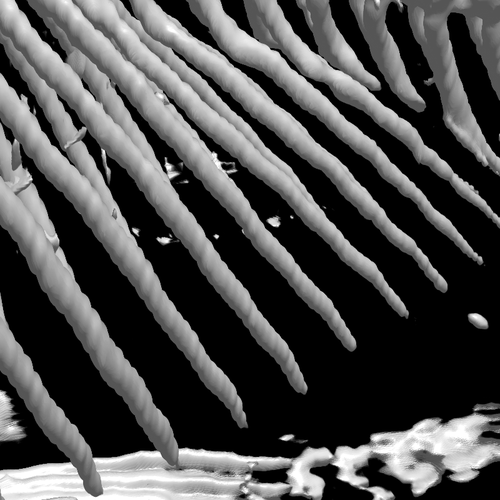}
    \end{subfigure}%
    \\
    \begin{subfigure}{0.242\textwidth}
        \includegraphics[width=\linewidth]{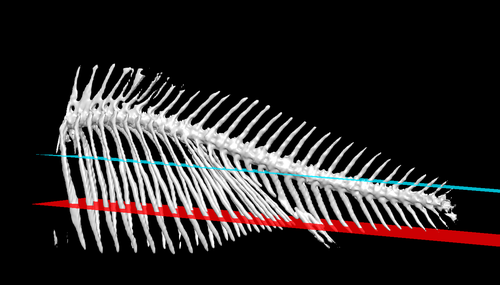}
    \end{subfigure}%
    \begin{subfigure}{0.379\textwidth}
        \includegraphics[width=\linewidth]{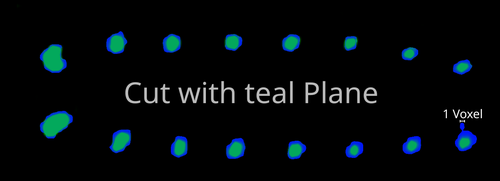}
    \end{subfigure}%
    \begin{subfigure}{0.379\textwidth}
        \includegraphics[width=\linewidth]{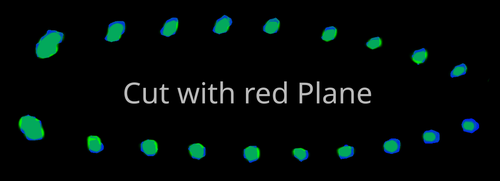}
    \end{subfigure}%
    
    \caption{\textbf{Segmentation of Fine Details.} Here we assess the quality of segmentation for fine details using the 3D bilateral solver. The \textbf{top row} shows our approach applied to the fine fish bones of the \textsc{Carp} dataset. The \textbf{middle row} shows a volume rendering using a 1D transfer function to visualize the fish bones. The 1D TF should - besides the presence of additional structures - fully reveal the fish bones. The zoomed view on the right shows the region that is marked orange in the top center. The \textbf{bottom row} compares the cross sections of the fine fish bones with the teal and red planes marked on the left. The intersection with teal can be seen in the center and with the red plane on the right. In those intersection images, \textbf{green} shows the border found by the 1D TF and \textbf{blue} shows what Ours+BLS finds.}
    \label{fig:fine-details}
\end{figure*}

In our approach we use the 3D bilateral solver to (1) increase the resolution of the segmentation and (2) to adapt the coarse segmentations to exact borders in the raw input volume. As very fine structures cannot be properly segmented in our initial similarity map $\ntfSL$ due to lack of resolution, the bilateral solver needs to restore the fine details in the refinement step. In Figure~\ref{fig:fine-details} we present a quality assessment of this process, by showing segmentations of fine fish bones in the \textsc{Carp} dataset. For this we compare our segmentation with a carefully crafted 1D transfer function. Due to the nature of the CT scan, the fish bones can be segmented fairly well using just the 1D transfer function, albeit with some additional noise and unwanted structures. This 1D TF segmentation of the fish bones serves as ground truth for our comparison. Figure~\ref{fig:fine-details} shows our approach on top, the 1D TF in the middle row and a comparison of two cross sections of those fish bones in the bottom row. Generally our approach detects all fish bones well and manages to avoid the noise present when using a 1D transfer function. The bottom row indicates that our approach makes the fish bones appear slightly thicker than the 1D TF, but gets within 1 voxel distance for most bones.

\bibliographystyle{abbrv}
\bibliography{references}